\documentclass[nohyperref]{article}

\usepackage{graphicx}
\usepackage{caption}
\usepackage{subcaption}
\usepackage{booktabs}
\usepackage{color}
\usepackage{microtype}
\usepackage{hyperref}

\usepackage[accepted]{icml2022}
\usepackage{amsmath}
\usepackage{amssymb}
\usepackage{mathtools,nccmath}
\usepackage{amsthm}
\usepackage{smile}

\newcommand{\norm}[1]{\left\lVert#1\right\rVert}

\begin{document}
\icmltitlerunning{Wide Neural Networks Forget Less Catastrophically}

\twocolumn[
\icmltitle{Wide Neural Networks Forget Less Catastrophically}
\icmlsetsymbol{equal}{*}
\icmlsetsymbol{dagger}{$\dagger$}

\begin{icmlauthorlist}
\icmlauthor{Seyed Iman Mirzadeh}{equal,wsu}
\icmlauthor{Arslan	Chaudhry}{goog,dagger}
\icmlauthor{Dong	 Yin}{goog}\\
\icmlauthor{Huiyi	 Hu}{goog}
\icmlauthor{Razvan Pascanu}{goog}
\icmlauthor{Dilan	Gorur}{goog}
\icmlauthor{Mehrdad Farajtabar}{goog,dagger}
\end{icmlauthorlist}

\icmlaffiliation{goog}{DeepMind}
\icmlaffiliation{wsu}{Washington State University}

\icmlcorrespondingauthor{Seyed Iman Mirzadeh}{seyediman.mirzadeh@wsu.edu}
\icmlcorrespondingauthor{Mehrdad Farajtabar}{farajtabar@google.com}

\icmlkeywords{Machine Learning, ICML, Continual Learning, Neural Network, Lifelong Learning, Catastrophic Forgetting, Wide Neural Networks}

\vskip 0.3in
]
\printAffiliationsAndNotice{\icmlEqualContribution} %

\begin{abstract}
A primary focus area in continual learning research is alleviating the ``catastrophic forgetting'' problem in neural networks by designing new algorithms that are more robust to the distribution shifts. While the recent progress in continual learning literature is encouraging, our understanding of what properties of neural networks contribute to catastrophic forgetting is still limited. To address this, instead of focusing on continual learning algorithms, in this work, we focus on the model itself and study the impact of ``width'' of the neural network architecture on catastrophic forgetting, and show that width has a surprisingly significant effect on forgetting. To explain this effect, we study the learning dynamics of the network from various perspectives such as gradient orthogonality, sparsity, and lazy training regime. We provide potential explanations that are consistent with the empirical results across different architectures and continual learning benchmarks.

\end{abstract}

\section{Introduction}
Machine learning is relying more and more on training large models on large static datasets to reach impressive results~\citep{kaplan2020scaling,lazaridou2021pitfalls,hombaiah2021dynamic}.
However, the real world is changing over time, and new information is becoming available at an unprecedented rate~\citep{lazaridou2021pitfalls,hombaiah2021dynamic}. In such real-world problems, the learning agent is exposed to a continuous stream of data, with potentially changing data distribution, and it has to absorb new information efficiently while not being able to iterate on previous data as freely as wanted due to time, sample, compute, privacy, or environmental complexity issues~\citep{Parisi2018ContinualLL}. To overcome these inefficiencies, fields, such as \emph{continual learning} (CL)~\citep{ring1994continual} or lifelong learning~\citep{thrun1995lifelong} are gaining a lot of attention recently. One of the key challenges in continual learning models is the abrupt erasure of previous knowledge, referred to as \emph{catastrophic forgetting} (CF)~\citep{McCloskey1989CatastrophicII}.

 Alleviating catastrophic forgetting has attracted a lot of attention lately, and many interesting solutions are proposed to partly overcome the issue~\citep[e.g.,][]{toneva2018empirical,nguyen2019toward, hsu2018re,li2019learn,Wallingford2020InTW}. These solutions vary in degree of complexity from simple replay-based methods to complicated regularization or network expansion-based methods. Unfortunately, however, there is not much fundamental understanding of the intrinsic properties of neural networks that affect continual learning performance through catastrophic forgetting or forward/backward transfer~\citep{mirzadeh2020understanding}. Moreover, most of the continual learning works focus on the algorithmic side of alleviating the forgetting rather than the architecture being used throughout the learning experience. This motivated our work to study one of the most important aspects of neural network architectures: the structure of the neural networks (e.g., width and depth).

 In contrast to the orthodox importance of the depth for network performance, such as in AlexNet~\citep{krizhevsky2012imagenet}, Inception~\citep{szegedy2016rethinking}, ResNets~\citep{he2016deep} or GPT family of models~\citep{brown2020language}, we study how the width of the network affects the continual learning performance -- aligned with the recent research on infinite width networks and the neural tangent kernel (NTK) regime~\citep{jacot2018neural,lee2019wide,arora2019exact,chizat2019lazy}. 
 We empirically demonstrate that increasing the width alone reduces catastrophic forgetting significantly, while it's not the case for depth.
 
\begin{figure}[t]
\centering
\begin{subfigure}{.48\linewidth}
    \centering
    \includegraphics[width=00.99\textwidth]{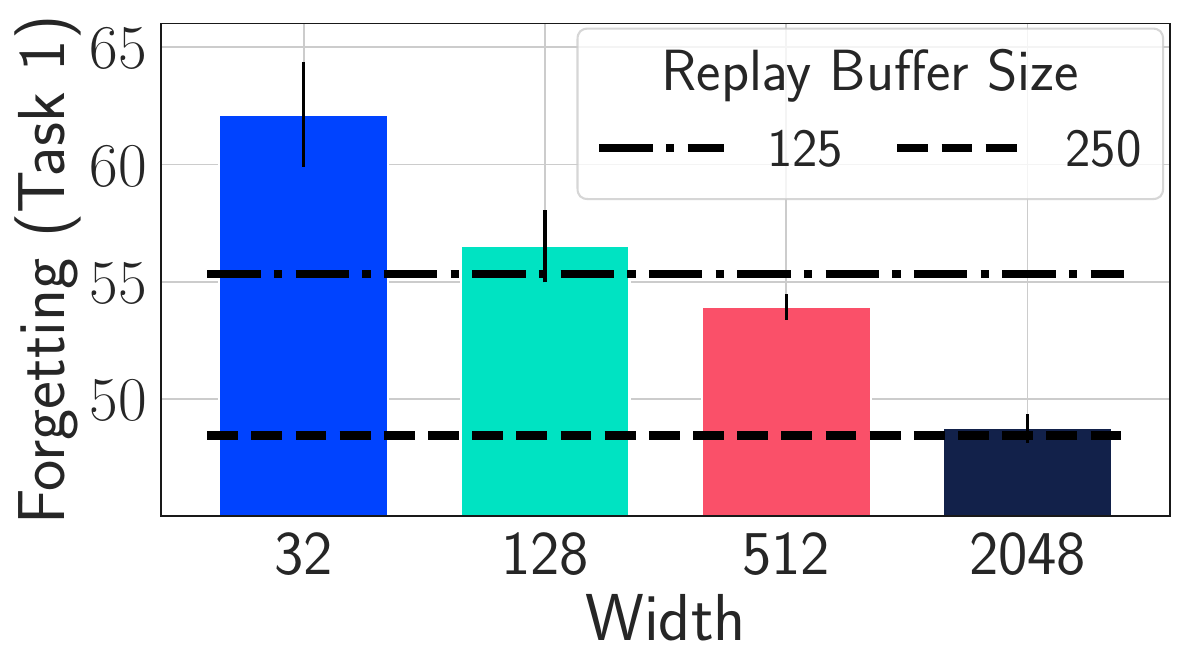}
    \caption{MLP on Rotated MNIST }
    \label{fig:intro-width-mnist}
\end{subfigure}\hfill
\begin{subfigure}{.48\linewidth}
    \centering
    \includegraphics[width=00.99\textwidth]{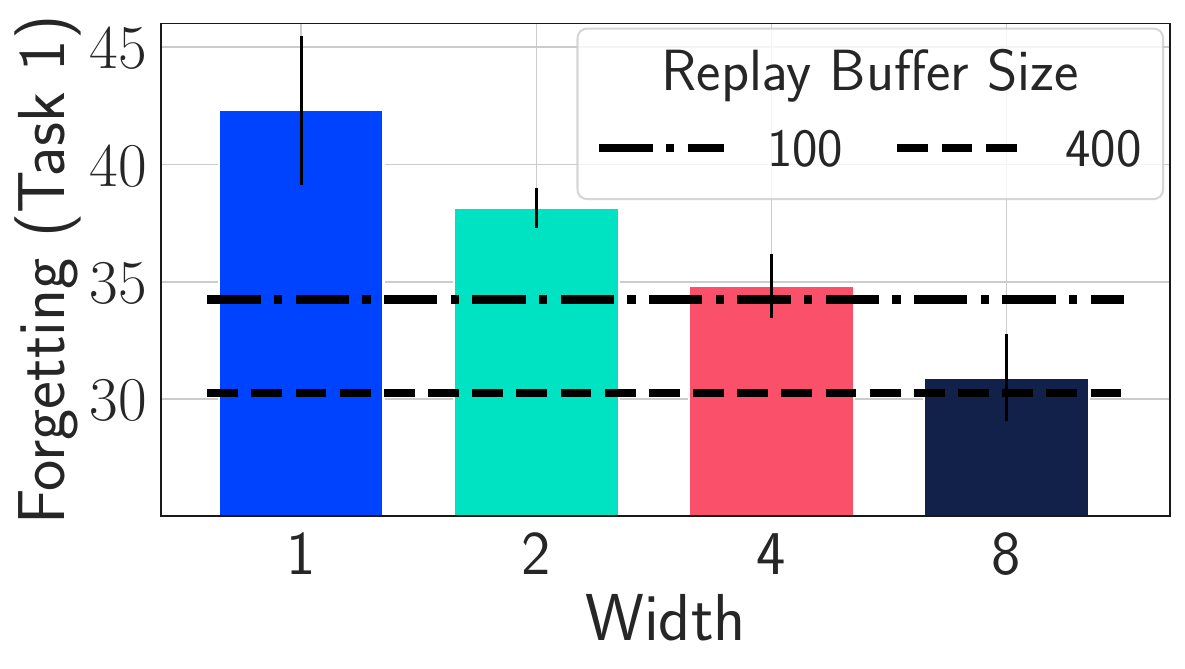}
    \caption{WRN on Split CIFAR-100 }
    \label{fig:intro-width-cifar}
\end{subfigure}
\hfill
\caption{Amount of forgetting w.r.t. increasing width. Wider networks forget less.}
\label{fig:intro-width}
\end{figure}

\begin{figure}[t]
\centering
\begin{subfigure}{.48\linewidth}
    \centering
    \includegraphics[width=0.99\textwidth]{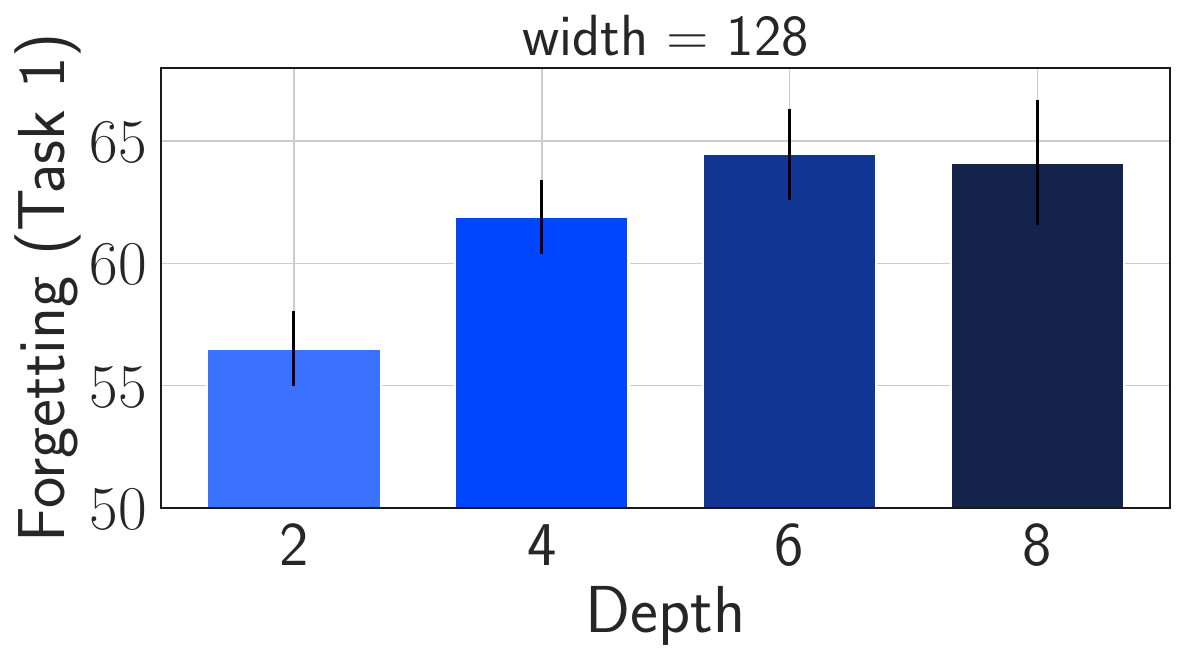}
    \caption{MLP on Rotated MNIST }
    \label{fig:intro-depth-mnist}
\end{subfigure}\hfill
\begin{subfigure}{.48\linewidth}
    \centering
    \includegraphics[width=0.99\textwidth]{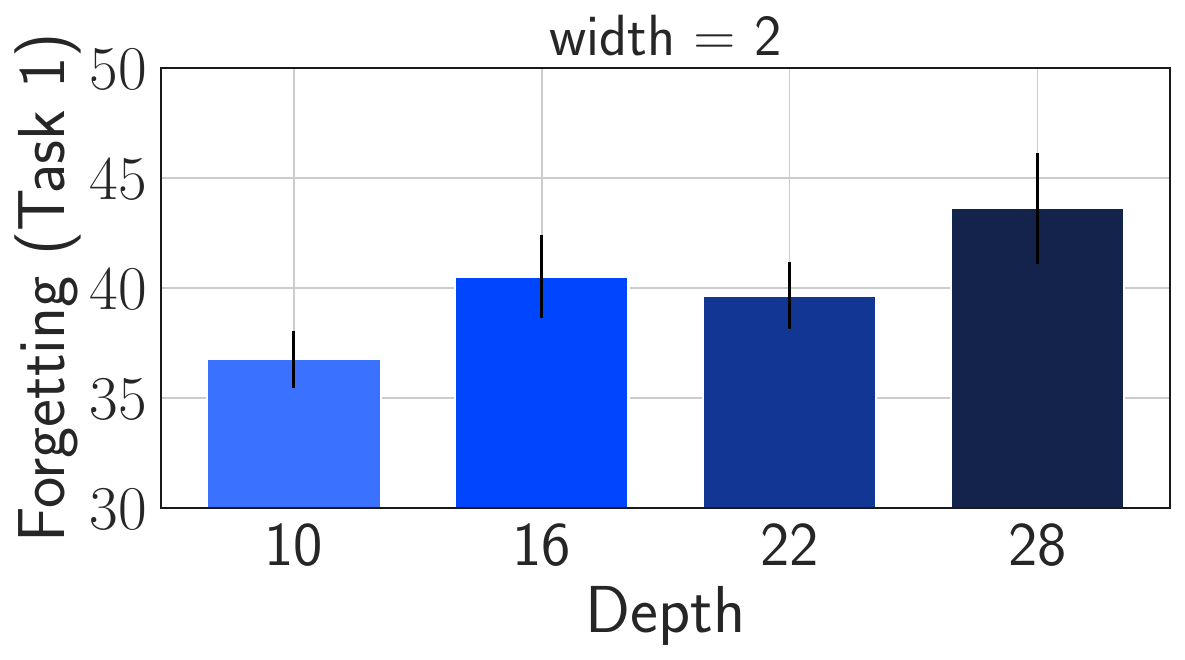}
    \caption{WRN on Split CIFAR-100 }
    \label{fig:intro-depth-cifar}
\end{subfigure}
\caption{Amount of forgetting w.r.t. increasing depth.}
\label{fig:intro-depth}
\end{figure}

 Concretely, Figures~\ref{fig:intro-width} and~\ref{fig:intro-depth} depict our findings on two popular continual learning benchmarks. Figure~\ref{fig:intro-width}
 shows that by increasing the width of a 2-layer MLP from 32 to 2048 the amount of forgetting of task 1 (after learning the 5th task) decreases from 62 to 48 percent.
 On Split CIFAR-100, by multiplying the width of a WideResNet-10 by 8 we can decrease forgetting of the first task (after learning the 20th task) from 42 to 31 percent.

  The boost in performance gained by only increasing the width is comparable to employing alternative methods specially developed to tackle CF with moderately sized replay buffers or regularization schemes. For example, it can be seen that in Figure~\ref{fig:intro-width-mnist}, wider networks without replay achieve comparable or better performance than the network with width 32 and employing replay buffers of sizes of 125 and 250. A similar level of performance improvement can be observed in Figure~\ref{fig:intro-width-cifar}, where the baseline WideResNet (width factor equal to 1) is trained with replay buffer sizes of 100 and 400. The ability of the wider networks to be less forgetful points towards looking at forgetting through the lens of over-parametrization~\citep{kaplan2020scaling,lazaridou2021pitfalls,hombaiah2021dynamic}.
  
  It is noteworthy that increasing width is not aimed to achieve state-of-the-art performance in alleviating forgetting. Our goal is not to compete with computationally cheaper and performant rehearsal or regularization-based methods~\citep{chaudhry2018efficient,farajtabar2020orthogonal,mirzadeh2021linear} but to provide a perspective on how over-parametrization and especially the width of the neural network can affect CF. One may wonder that a similar over-parametrization achieved through depth might have a similar positive effect on continual learning performance. Figure~\ref{fig:intro-depth} shows that this is not the case and that achieving over-parametrization through depth has no or even negative effect on mitigating forgetting. 
  
 \textbf{Contributions.} We study the implications of different neural network structures and over-parametrization (i.e., width and depth) for continual learning. We show that increased width can alleviate the forgetting and provide simplistic theoretical and empirical justifications such as lazy training regime and orthogonalization that can explain the benefits of using wide neural networks in continual learning.

 \section{Related Work} \label{sec:related_work}
In continual learning~\citep{ring1994continual}, also referred to as lifelong learning~\citep{thrun1995lifelong}, an agent, constrained in time and memory, receives a sequence of tasks. The goal of the agent is to remember the useful knowledge of the past tasks to solve new tasks efficiently. Catastrophic forgetting~\citep{McCloskey1989CatastrophicII,mcclelland1995there,goodfellow2013empirical}---a sudden erasure of previous knowledge when exposed to new tasks---is identified as one of the key challenges in continual learning. Recently, a lot of algorithmic progress has been made in mitigating catastrophic forgetting in neural networks. The progress can broadly be classified into three categories. 

\textbf{Regularization-based} methods~\citep{Rebuffi2016iCaRLIC,EWC,zenke2017continual,nguyen2017variational,aljundi2018memory,yin2020optimization} attempt to identify important parameters or features of the past tasks, and modify the training objective of the new task such that those parameters or features do not drift significantly. While regularization-based methods provide strong results for a small number of tasks, the importance of parameters drifts when the number of tasks is large~\citep{chaudhry2018efficient,titsias2019functional}. 

\textbf{Expansion-based} methods~\citep{rusu2016progressive,fernando2017pathnet,aljundi2017expert,rosenbaum2018routing,chang2018,rcl2018,modularmetal2018,li2019learn,veniat2020efficient} add new network modules or experts for new tasks. By construction, expansion-based methods can have zero-forgetting, but their memory complexity can increase super-linearly with the number of tasks, making them undesirable when the pool of tasks is large. We note that even though the expansion-based works may have the model as their main subject, the main objective of these works is designing algorithms that can efficiently use a model (e.g., if the model should be expanded or not), and they do not explicitly analyze the structure of the model (e.g., width). For instance, while \citet{yoon2018lifelong} show that expanding the network capacity thorough width is helpful, they have not studied the impact of increasing capacity when the depth increases, nor why increasing the width is helpful. Overall, in this work, we are interested in understanding the impacts of network structure (e.g., width and depth) on continual learning performance by focusing on the model itself rather than the algorithm that works with the model.

Perhaps the strongest are the \textbf{replay-based} methods that store a small subset of data from previous tasks in the replay buffer, and either use it directly in the form of experience replay with new task~\citep{shin2017continual,kirichenko2021task,rolnick2018experience,riemer2018learning,Chaudhry2019OnTE,chaudhryOrthog2020,balaji2020effectiveness} or use it as a set of optimization constraints~\citep{lopez2017gradient,chaudhry2018efficient,farajtabar2020orthogonal}. 

\looseness=-1 While these methods provide algorithmic improvements on top of standard neural network training (empirical risk minimization), there is still limited understanding as to what causes forgetting in the first place. Some works have looked at the forgetting from the lens of network representations. For example, \citet{javed2019meta,beaulieu2020learning} showed that meta-learned representations forget less when trained continually. On the other hand, \citet{SanketPreTrain21} showed that pre-trained over-parameterized networks forget considerably less than randomly initialized networks when trained continually. Similarly, few works have studied the forgetting from the neural network's training dynamics perspective. \citet{mirzadeh2021linear} showed that the optima of the multitask learning and continual learning solutions are linearly connected in the parameter space. Contrary to these works, we look at catastrophic forgetting from the lens of network architecture, specifically with respect to its width and depth and show that wider networks forget considerably less. 

A flurry of recent papers in deep learning theory has taken a huge step in understanding the training behavior of neural networks. In particular, \citet{jacot2018neural} first introduced the concept of  neural tangent kernel (NTK) by studying the training dynamics of neural networks with square loss in the infinite width or ``over-parameterized'' regime. The resulting neural network can be approximated by a linear model.  Based on the NTK framework, other works \citep{allen2019learning,du2019gradient,zou2020gradient} have proved the convergence of the SGD on various other settings and proved the NTK to be a powerful framework to study neural networks training dynamics.
Later~\citet{chizat2019lazy} developed a parallel idea under the framework called ``lazy training'' where the weights stay near their initial values during the training. In this regime, the system becomes almost linear, and the dynamics of SGD within this region can be characterized via properties of the associated NTK. In the context of continual learning,~\citet{doan2021a} have studied forgetting of projection-based algorithms in the NTK regime using the similarity of the tasks.

Finally, several recent works have explored the intersection of neural network architectures and continual learning. \citet{Mirzadeh2022ArchitectureMI} provides a comprehensive analysis on the significant role of architectures and architectural decisions in continual learning. They show that that different architecture families have different learning and retention capabilities, and compare the merits and flaws of various architectures. For the pre-training setup, \citet{EffectOfScaleInCL} studies the effect of scaling in continual learning and show that large pre-trained models perform significantly better than the models trained from scratch in continual learning scenarios.

\section{Main Experiments} \label{sec:experiments}
In this section, we study the impact of width on both average accuracy and average forgetting. Further detailed experiments and results will be provided in Sections~\ref{sec:analysis} and~\ref{sec:additional-results}.

\subsection{Experimental Setup}
The experimental setup, such as benchmarks, network architectures, continual learning setting (e.g., number of tasks, episodic memory size, and training epochs per task), hyperparameters, and evaluation metrics are chosen to be similar to several other studies~\citep{Chaudhry2019OnTE,farajtabar2020orthogonal,mirzadeh2021linear}. To ensure that the reported results are not more favorable to a specific architecture, we use a grid of hyper-parameters, and for each model, we report the results with the best hyperparameters. For all experiments, we report the average and standard deviation over five runs with different random seeds for network initialization. In addition to random seed for initialization, for Split CIFAR-100, we use three different seeds for each run where each seed corresponds to the random order in which classes are selected for each task. We provide more details about our experimental design in Appendix~\ref{apx:experimental-setup}.

{\bf Benchmarks.}
We report our findings on two standard continual learning benchmarks: Rotated MNIST and Split CIFAR-100. In Rotated MNIST, each task is generated by the continual rotation of the MNIST images for degrees 0, 22.5, 45, 67.5, and 90, respectively, constituting 5 different tasks. For Split CIFAR-100 each task contains the data from 5 random classes (without replacement), resulting in 20 tasks. Finally, in both benchmarks, each model trains on each task for five epochs. We have selected Rotated MNIST as a domain-incremental benchmark and Split CIFAR-100 as a task-incremental benchmark to represent two common scenarios in continual learning.

{\bf Networks.}
For the Rotated MNIST benchmark, we use a two-layer MLP with ReLU non-linearities in each layer and vary the width. For the Split CIFAR-100 benchmark, we use a WideResNet (WRN) with the depth of 10~\citep{zagoruyko2016wide} and different width factors which multiply the width of the layers (i.e., the number of convolutional channels) in the original WideResNet by $\{1, 2, 4, 8\}$. All models are trained using the SGD optimizer, and the best result for each model is reported. %

{\bf Evaluations.}
Following previous work~\citep{chaudhry2018efficient,riemer2018learning,mirzadeh2021linear}, we report the following metrics:\\
(1) \emph{Average Accuracy}: The average validation accuracy after the model has been continually trained for $T$ tasks, is defined as:
\begin{equation}\label{eq:avg_acc}
    A_T= \frac{1}{T} \sum_{i=1}^T a_{T,i}
\end{equation}
where, $a_{t,i}$ is the validation accuracy on the dataset of task $i$ after the model finished learning task $t$.\\
(2) \emph{Average Forgetting}: The average forgetting is calculated as the difference between the peak accuracy and the final accuracy of each task, after the continual learning experience is finished. For a continual learning benchmark with $T$ tasks, it is defined as:

\begin{equation}\label{eq:avg_forg}
    F = \frac{1}{T-1} \sum_{i=1}^{T-1}{\text{max}_{t \in \{1,\dots, T-1\}}~(a_{t,i}-a_{T,i})}
\end{equation}

(3) \emph{Learning Accuracy}: The accuracy for each task directly after it is learned. The learning accuracy provides a good representation of the plasticity of a model and can be calculated using: 
\begin{equation}\label{eq:learning_acc}
    \text{LA}_T = \frac{1}{T} \sum_{i=1}^T {a_{i, i}}
\end{equation}

(4) \emph{Joint Accuracy}: The accuracy of the model trained on the combined data of all tasks together.

\subsection{Main Results}
Tables~\ref{tab:results-mnist} and \ref{tab:results-cifar} show the main results in terms of average accuracy, average forgetting, and learning accuracy for MLP on Rotated MNIST and WideResent on Split CIFAR-100 with varying widths. It can be seen from the tables that wider networks improve both average accuracy and forgetting. 

It is important to note that reduced forgetting is not due to stabilizing the model dogmatically at the cost of plasticity. Increased average accuracy means that the wider models not only remember previous tasks better, but keep their competitive performance on the current task too. This can be further consolidated by observing the learning accuracy (LA) of the wider networks.

\begin{table}[t]
\centering
\caption{MLP on Rotated MNIST}
\label{tab:results-mnist}
\resizebox{\linewidth}{!}{%
\begin{tabular}{@{}ccccc@{}}
\toprule
\textbf{Width} & \textbf{\begin{tabular}[c]{@{}c@{}}Average\\ Accuracy\end{tabular}} & \textbf{\begin{tabular}[c]{@{}c@{}}Average\\  Forgetting\end{tabular}} & \textbf{\begin{tabular}[c]{@{}c@{}}Learning \\ Accuracy\end{tabular}} & \textbf{\begin{tabular}[c]{@{}c@{}}Joint\\ Accuracy\end{tabular}} \\ \midrule
32 & 65.9 $\pm 1.00 $ & 36.9 $\pm 1.27$ & 95.5 $\pm 0.85$ & 91.2 $\pm$0.57 \\
128 & 70.8 $\pm 0.68$ & 31.5 $\pm 0.92$ & 96.0 $\pm 0.90$ & 93.4 $\pm$0.66 \\
512 & 72.6 $\pm 0.27$ & 29.6 $\pm 0.36$ & 96.4 $\pm 0.73$ & 94.1 $\pm$0.77 \\
2048 & 75.2 $\pm 0.34$ & 26.7 $\pm 0.50$ & 96.6 $\pm 0.61$ & 94.0 $\pm$0.45 \\ \bottomrule
\end{tabular}%
}
\end{table}

\begin{table}[t]
\centering
\caption{WideResNet-10 on Split CIFAR-100}
\label{tab:results-cifar}
\resizebox{\linewidth}{!}{%
\begin{tabular}{@{}ccccc@{}}
\toprule
\textbf{Width} & \textbf{\begin{tabular}[c]{@{}c@{}}Average\\ Accuracy\end{tabular}} & \textbf{\begin{tabular}[c]{@{}c@{}}Average\\ Forgetting\end{tabular}} & \textbf{\begin{tabular}[c]{@{}c@{}}Learning\\ Accuracy\end{tabular}} & \textbf{\begin{tabular}[c]{@{}c@{}}Joint\\ Accuracy\end{tabular}} \\ \midrule
1 & 47.1 $\pm 2.60$ & 37.3 $\pm 2.62$ & 81.1 $\pm 4.37$ & 81.4 $\pm 0.62$ \\
2 & 49.7 $\pm 1.51$ & 34.9 $\pm 1.72$ & 82.9 $\pm 5.17$ & 83.6 $\pm 0.13$ \\
4 & 53.8 $\pm 2.74$ & 33.8  $\pm 2.16$ & 83.9 $\pm 6.53$ & 84.4 $\pm 0.64$ \\
8 & 59.7 $\pm 2.33$ & 29.4 $\pm 2.52$ & 87.5 $\pm 4.15$ & 84.8 $\pm 0.49$ \\ \bottomrule
\end{tabular}%
}
\end{table}

\section{Analysis}\label{sec:analysis}
\looseness=-1In this section, we overview the reasons we believe contribute to the reduction in catastrophic forgetting when widening the network.
Although wide networks have higher capacity, similar to how it has been shown that wider networks generalize well because of implicit regularization imposed by gradient descent~\citep{jacot2018neural,lee2019wide}, we argue that the behavior of catastrophic forgetting cannot be reduced to capacity alone and we need to integrate learning dynamics.
Here, we first construct a simple example  for a regression problem with squared loss to demonstrate this point.
Informally, we have the following claim.

\begin{claim}[informal]\label{claim:fgt}
Consider learning problems with input space $\RR^d$, output space $\RR$, and squared loss. Let $\cF_1$ be the class of linear models that maps the input to the output and $\cF_2$ be the class of two-layer \textbf{linear} networks (i.e., no nonlinear activation). Then, there exist two tasks such that when we train task $2$ using gradient descent, if we use model class $\cF_1$, the amount of forgetting for task $1$ is strictly zero;
whereas if we use model class $\cF_2$, the amount of forgetting can be positive.
\end{claim}

We provide details of the construction of the two tasks in Appendix~\ref{apx:task_construction}. As we can see, the two model classes $\cF_1$ and $\cF_2$ have exactly the same capacity, i.e., $\cF_1=\cF_2$, since there is no nonlinearity in $\cF_2$. However, due to the use of gradient descent algorithm, the amount of forgetting for task $1$ is different for the two model classes when training on task $2$. Moreover, in Appendix~\ref{apx:task_construction}, we show that if the size of a hidden layer in $\cF_2$ is not large enough, then the forgetting could be more severe. Intuitively, if the additional layer becomes a bottleneck that projects the input to a very low dimensional space, the information can be lost.

While this example is a specially constructed one, we believe it can be helpful towards understanding the benefits of wide neural networks in continual learning. In particular, it shows that we have to take the commonly used \emph{gradient descent algorithm} into account when investigating this phenomenon. In Appendix~\ref{sec:appendix-additional-results}, we provide further empirical results regarding the significance of the width of the first layer. In the following, we dive deeper in this phenomenon and provide several potential explanations.

\begin{figure*}[t]
\centering
\begin{subfigure}{.49\linewidth}
      \centering
      \includegraphics[width=.99\linewidth]{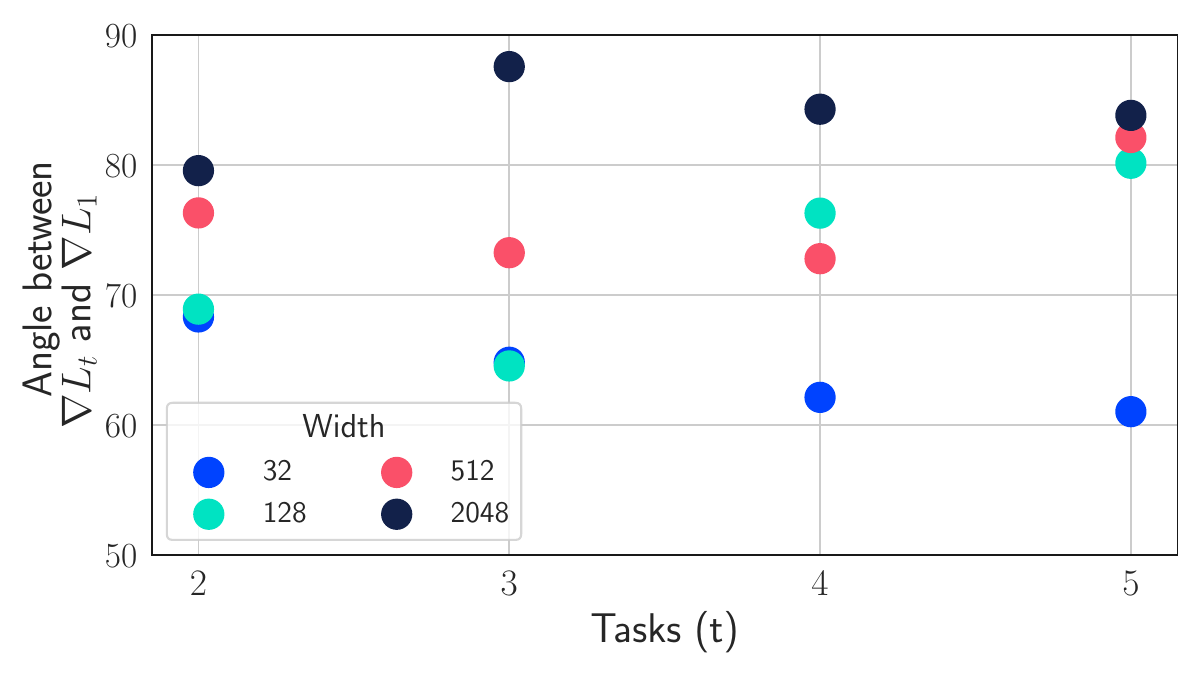}
      \caption{MLP on Rotated MNIST}
      \label{fig:ortho-mnist}
\end{subfigure}\hfill
\begin{subfigure}{.49\linewidth}
      \centering
      \includegraphics[width=.99\linewidth]{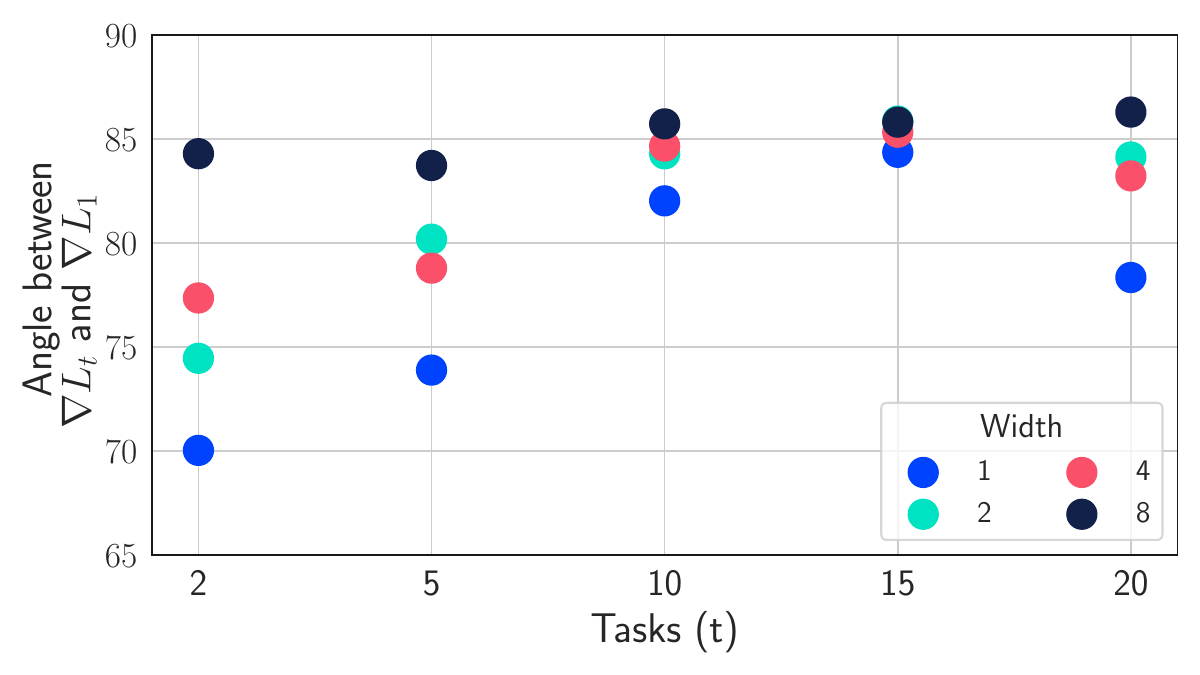}
      \caption{WRN on Split CIFAR-100}
      \label{fig:ortho-cifar}
\end{subfigure}\hfill
\caption{The angle between the gradients at the optimum of task 1, $\nabla L_1 (w_1^*)$, and between the optimum of subsequent tasks $t$, $\nabla L_t (w_t^*)$, for $t > 1$. Wider networks remain closer to 90 degree (more orthogonal).
}
\label{fig:ortho}
\end{figure*}

\subsection{Increased Gradient Orthogonality}

Earlier works in continual learning~\citep{farajtabar2020orthogonal,chaudhry2018efficient,lopez2017gradient,riemer2018learning} have focused on explicitly forcing the gradients of the new tasks to be orthogonal to those of the previous ones to reduce forgetting among the tasks. Here, we want to look at how the gradient directions of different tasks evolve as the continual learning experience progresses while the networks' width increases. We first establish why orthogonalizing the gradients of different tasks is a good continual learning strategy.

Let us consider a two-task continual learning problem, and study the amount of forgetting for task 1 induced by one gradient descent step on task 2. Let $\cW\subseteq\RR^p$ be the model parameter space and $L_t(w) : \cW \mapsto \RR$ be the training loss function of the $t$-th task. Assuming that all the loss functions are differentiable, according to mean value theorem, we have the following simple observation.
\begin{claim}\label{claim:one_step}
Let $w'=w - \eta \nabla L_2(w)$. Then there exists $\xi\in[0, 1]$ such that 
\begin{align}\label{eq:one_step_forget}
   L_1(w')-L_1(w) = -\eta\langle \nabla L_1(w-\xi\eta\nabla L_2(w)), \nabla L_2(w)\rangle. 
\end{align}
\end{claim}
As we can see, the left hand side of Eq.~\ref{eq:one_step_forget} is the increase in the loss of task $1$, and the right hand side is related to the inner product of the gradient of $L_1$ at $w-\xi\eta\nabla L_2(w)$, and the gradient of $L_2$ at $w$. Therefore, if the gradients of the two tasks are more \emph{orthogonal} to each other, then when training on new tasks, the model forgets less about previous task. Formally, we have the following direct corollary.
\begin{claim}\label{claim:ortho}
Suppose that for any $w_1,w_2\in\cW$ with $\|w_1-w_2\|_2\le \eta \sup_{w\in\cW} \|\nabla L_2(w)\|_2$, we have
\[
|\langle \nabla L_1(w_1), \nabla L_2(w_2)\rangle|\le \epsilon /\eta,
\]
for some $\epsilon>0$, and let $w'=w-\eta\nabla L_2(w)$. Then, we have $L_1(w')-L_1(w) \le \epsilon$.
\end{claim}

Hence, the one-step forgetting is small if we have near-orthogonal gradients for the two tasks. The next question is whether widening the network increases the gradient orthogonality. Empirically, to study the directions of the gradients for different tasks, in Figure~\ref{fig:ortho}, we plot the angle between the gradient vector at the minimum of the first task and that of the subsequent tasks for various width factors. For this figure, we concatenate the gradients from all the layers into one vector. It can be seen from the figure that the gradients at the minimum of subsequent tasks become more orthogonal to the gradients at the minimum of the first task as we increase the width. This trend also holds when the gradients of other tasks are used as reference (\emph{i.e.}) the gradients at the minimum of tasks $>$ 2 also become orthogonal to the gradients at the minimum of task 2. Refer to the appendix for these results. Since the gradients of different tasks generally become more orthogonal for the wider networks, the reduction in forgetting can be attributed to this orthogonalization.

\subsection{Increased Gradient Sparsity}

\begin{figure*}[t]
\centering
\begin{subfigure}{.49\linewidth}
      \centering
      \includegraphics[width=.99\linewidth]{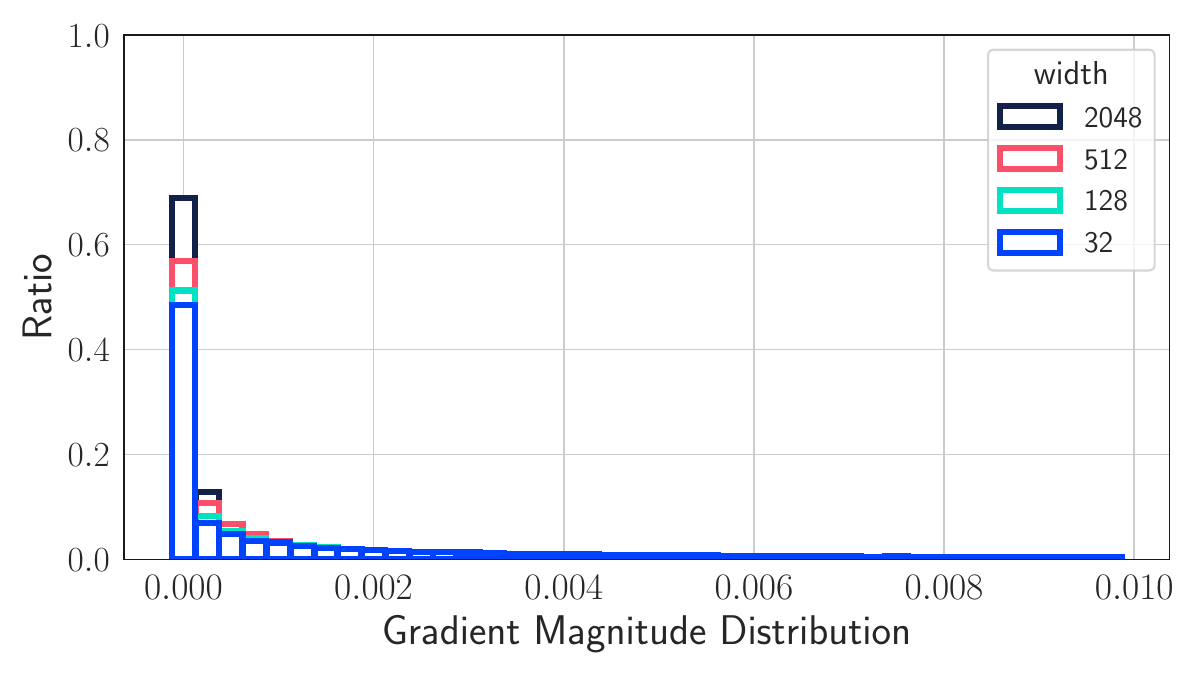}
      \caption{MLP on Rotated MNIST}
      \label{fig:sparse-mnist}
\end{subfigure}\hfill
\begin{subfigure}{.49\linewidth}
      \centering
      \includegraphics[width=.99\linewidth]{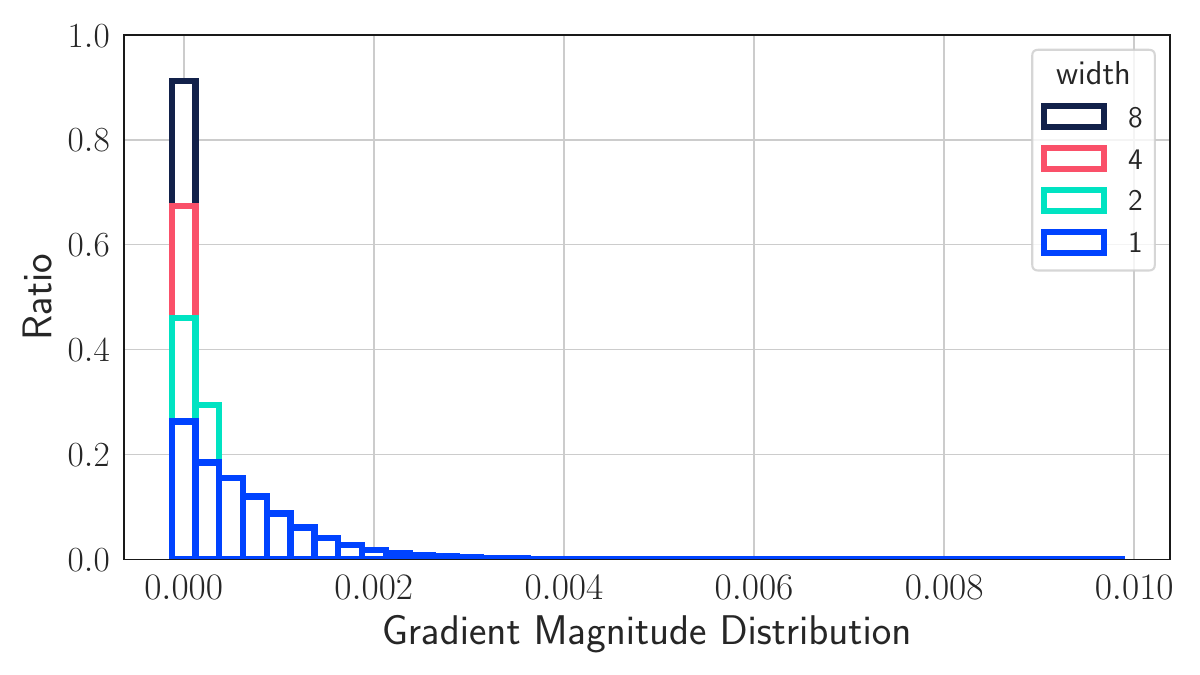}
      \caption{WRN on Split CIFAR-100 }
      \label{fig:sparse-cifar}
\end{subfigure}\hfill
\caption{Histogram of the magnitude of the elements of the gradient vector. Wider networks have sparser gradients (the histogram is more skewed towards the left).}
\label{fig:sparse}
\end{figure*}

Another scenario where forgetting may be reduced is that during the training of a new task, we only need to update a small number of model parameters, and thus the network is stable and forgets less about the previous tasks. In fact, according to Claim~\ref{claim:one_step}, we can establish the following simple corollary.
\begin{claim}\label{claim:sparse}
Suppose that $\|\nabla L_2(w)\|_2 \le B$ for all $w\in\cW$, and let $b(w):=\sup_{\|\tilde{w}-w\|_2\le \eta B}\|\nabla L_1(\tilde{w})\|_\infty$. Let $w' = w - \eta \nabla L_2(w)$. Then we have
\[
L_1(w')-L_1(w) \le \eta b(w)\|\nabla L_2(w)\|_0,
\]
where $\|\cdot\|_0$ denotes the number of non-zero elements in a vector.
\end{claim}

Therefore, increased gradient sparsity may lead to less forgetting.
Figure~\ref{fig:sparse} confirms this hypothesis by demonstrating the histogram of the absolute value of the entries of the gradient. The gradient norm is constructed by calculating the gradients of all layers on the empirical loss of task 2, after learning task 1 and before starting task 2. From the figure, it can be seen that the wider networks have sparser gradients. This implies a smaller number of parameters needs to change to adapt to new tasks resulting in a reduced forgetting on already learned tasks.

\subsection{Lazy Training Regime}\label{sec:lazy-training-regime}

In the previous two sections, we showed that the amount of forgetting induced by a single gradient descent step on a new task can be reduced by increased gradient orthogonality/sparsity.
In this section, we take a different perspective and study the entire training process of a task. In the following, we denote by $w^{\rm in}_t$ and $w^*_t$ the initial and end model parameters for task $t$, respectively. Note that in the continual learning scenario, the initializer for a subsequent task is in fact the solution to the previous task. Thus, $w^{\rm in}_t = w^{*}_{t-1}$.

Let us consider the amount of forgetting of task $t-1$ at the end of the training of task $t$, measured by the increase in loss function $L_{t-1}$, i.e., $L_{t-1}(w^*_t) - L_{t-1}(w^*_{t-1})$. Suppose that the loss function is differentiable, by mean value theorem, there exists $\xi\in[0, 1]$ such that
\begin{align*}
    &L_{t-1}(w^*_t) - L_{t-1}(w^*_{t-1}) \\
    =& \nabla L_{t-1}((1-\xi)w^*_{t-1}+\xi w^*_t)^\top (w^*_t - w^*_{t-1}),
\end{align*}
which yields the following upper bound on the amount of forgetting:
\begin{equation}\label{eq:dt_bound}
\begin{aligned}
& L_{t-1}(w^*_t) - L_{t-1}(w^*_{t-1}) \\
\le & D_t\sup_{\{w:\|w - w^*_{t-1}\|_2 \le D_t\}}\|\nabla L_{t-1}(w)\|_2,
\end{aligned}
\end{equation}
where $D_t :=\|w^*_t - w^*_{t-1}\|_2$ is the distance the the model parameter moves during the training of task $t$. This result suggests the following intuitive claim: The forgetting may be less severe if the model moves a shorter distance $D_t$ in the parameter space. 

\begin{figure*}[t]
\centering
\begin{subfigure}{.44\linewidth}
      \centering
      \includegraphics[width=.99\linewidth]{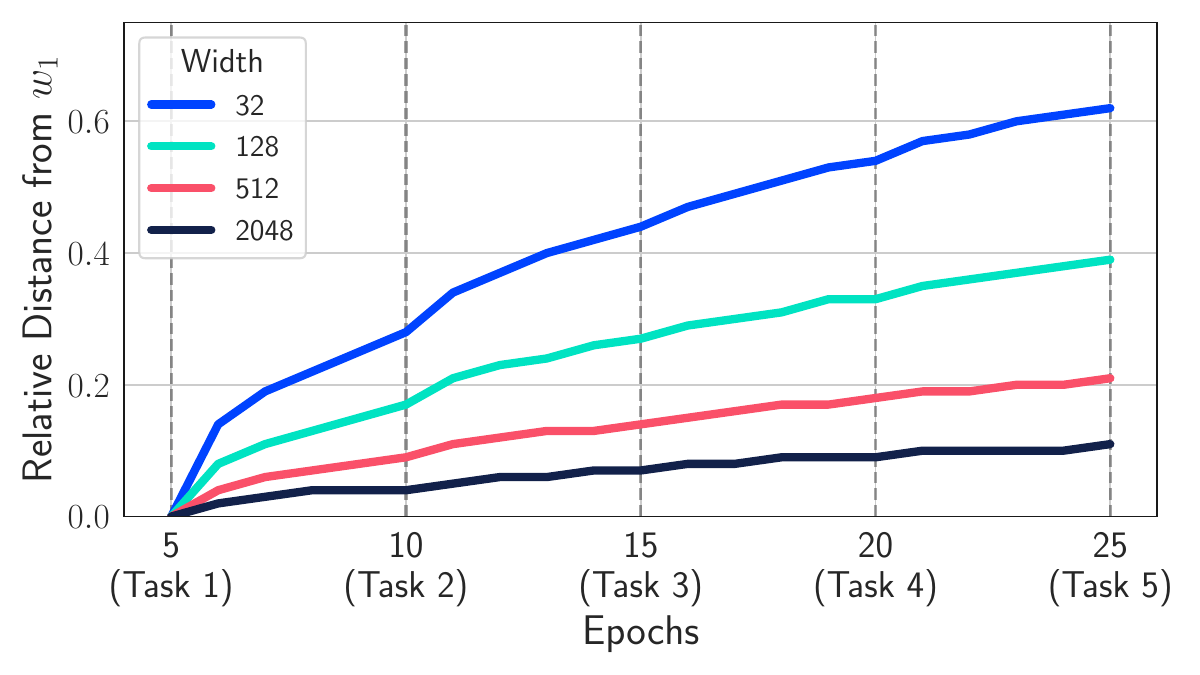}
      \caption{MLP on Rotated MNIST}
      \label{fig:lazy-mnist}
\end{subfigure}\hfill
\begin{subfigure}{.44\linewidth}
      \centering
      \includegraphics[width=.99\linewidth]{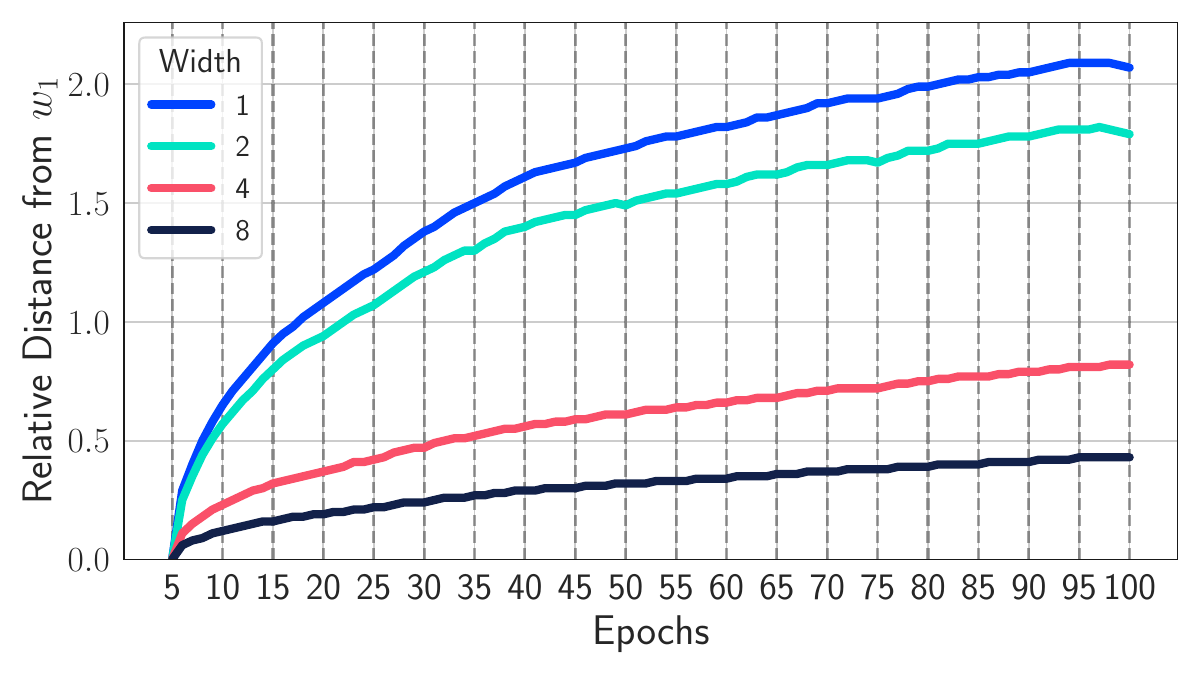}
      \caption{WRN on Split CIFAR-100}
      \label{fig:lazy-cifar}
\end{subfigure}\hfill
\caption{The distance between $w_1^*$, solution of task 1, and $w_t^*$, solution of the $t$-th task, for $t > 1$ over the course of training. Wider networks remain closer to the solution of the first task.}
\label{fig:lazy}
\end{figure*}

Interestingly, for overparametrized neural networks, the distance that the model parameter moves during SGD training can indeed be small. In their seminal papers, \citet{jacot2018neural} and ~\citet{lee2019wide} connect learning in neural networks to kernel methods and show that at the infinite width limit, the kernel doesn't change throughout the training, and the neural network evolves as a linear model under SGD training.
Moreover,~\citet{chizat2019lazy} show that these neural networks operate in lazy regime where $\| w^*-w^{\rm in} \|_2$
is very small. Since wider networks are closer to the infinite width limit, the moving distance $D_t$ may be smaller for wider networks and thus the forgetting is less severe. 

To confirm this intuition, in Figure~\ref{fig:lazy}, we plot the distance between the optimum of the first task and that of the subsequent tasks as the continual learning experience progress. In the figure, each task is trained for 5 epochs. From the figure, it can be seen that, for both MLPs and ResNets, the relative distance of the parameters for the wider networks is substantially smaller than that of the less wide networks. This confirms that lazy training of wider networks is indeed beneficial for continual learning and reduces forgetting significantly.
In the same vein, \citet{mirzadeh2020understanding, mirzadeh2020dropout} observed that changing the learning regime (i.e., imposing a learning rate decay, decreasing batch size, employing dropout, etc.) also leads to a small change in parameters from the optimum of previous tasks resulting in less forgetting. However, their work was concerned about the optimization properties of neural networks rather than architectural ones.

\subsection{The Case for Depth} \label{sec:depth_cf}

In Section~\ref{sec:experiments}, we saw that the width of the network has a beneficial effect when it comes to catastrophic forgetting. We do not see a similar monotonic reduction in forgetting when we increase the depth of the network in our experiments. In fact, we observe that with increasing depth, the forgetting increases, as shown in Figure~\ref{fig:intro-depth}. We now provide an intuitive analysis for why and when the network depth hurts the performance of a continual learning model.

Inspired by the exploding gradients phenomena in recurrent neural networks~\citep{bengio1994learning,pascanu2013difficulty}, we show that deeper networks may cause larger gradient updates in the earlier layers. Larger gradients mean that the solution for the subsequent tasks will not be found in the vicinity of the current task, which would entail increased forgetting. To establish the larger gradients for deeper networks, we use the standard argument~\citep{bengio1994learning} for exploding gradients. 
Suppose that the network consists of $K$ feed-forward layers parameterized by weight matrices $\{W_i\}_{i=1}^K$ and non-linearities $\sigma(\cdot)$. For ease of exposition, we ignore the biases and normalization layers in the network. The gradient on any intermediate layer $l$ can be written as:
\begin{align*}
\frac{\partial L}{\partial h_{l}} &= \frac{\partial L}{\partial h_K} . \frac{\partial h_K}{\partial h_l} = \frac{\partial L}{\partial h_K} . \prod_{i=l}^{K-1} \frac{\partial h_{i+1}}{\partial h_i} \\
&= \frac{\partial L}{\partial h_K}.\prod_{i=l}^{K-1} \Sigma_{i+1} W_{i+1},
\end{align*}
where $\Sigma_i$ is the diagonal Jacobian of the non-linearity at the $i$-th layer. Note, if a ReLU non-linearity is used, then $\norm{\Sigma_i} = 1$, unless all the activations in the layer are negative. 

The norm of the gradient on layer $l$ can then be bounded as:
\begin{align*}
\norm{\frac{\partial L}{\partial h_{l}}} &= \norm{\frac{\partial L}{\partial h_K}.\prod_{i=l}^{K-1} \Sigma_{i+1} W_{i+1}} \\
& \leq\norm{\frac{\partial L}{\partial h_K}}.\prod_{i=l}^{K-1}\norm{\Sigma_{i+1}} \norm{W_{i+1}} \\
& \le \norm{\frac{\partial L}{\partial h_K}}.\prod_{i=l}^{K-1}\norm{W_{i+1}}, 
\end{align*}
where in the second line, we use sub-ordinate and sub-multiplicative properties of induced matrix norms, and in the third line, we assumed that ReLU activation is used and thus $\norm{\Sigma_i} \le 1$. This analysis shows that the gradient norm on the earlier layers is proportional to the product of matrix norms of higher layer weights. For a 2-norm, the gradient magnitude could be bounded by the product of largest singular values ($\Lambda_i$) of the intermediate weight matrices:
\begin{equation}\label{eq:sv-bound}
\norm{\frac{\partial L}{\partial h_{l}}} \leq \norm{\frac{\partial L}{\partial h_K}}.\prod_{i=l}^{K-1}\norm{\Lambda_{i+1}}.
\end{equation}

\looseness=-4 Although, for $\Lambda_i > 1$ this bound in unusable, as it is an upper bound which grows exponentially, and becomes more and more loose, with $K$ (the number of layers) increasing, however, it does potentially point to a scenario when the gradients on the earlier layers could increase with the depth. This \emph{can} happen when the singular values of all the intermediate weight matrices are greater than 1, \emph{i.e.}, $\Lambda_i > 1, \forall i \in \{l, \cdots, K-1\}$. In the supplementary Figure~\ref{fig:sv}, we show that this indeed is the case for the datasets and architectures considered in this work. A large gradient on earlier layers can then contribute to a larger forgetting of previous tasks. 

\begin{figure*}[t]
\centering
\begin{subfigure}{.45\linewidth}
      \centering
      \includegraphics[width=.95\linewidth]{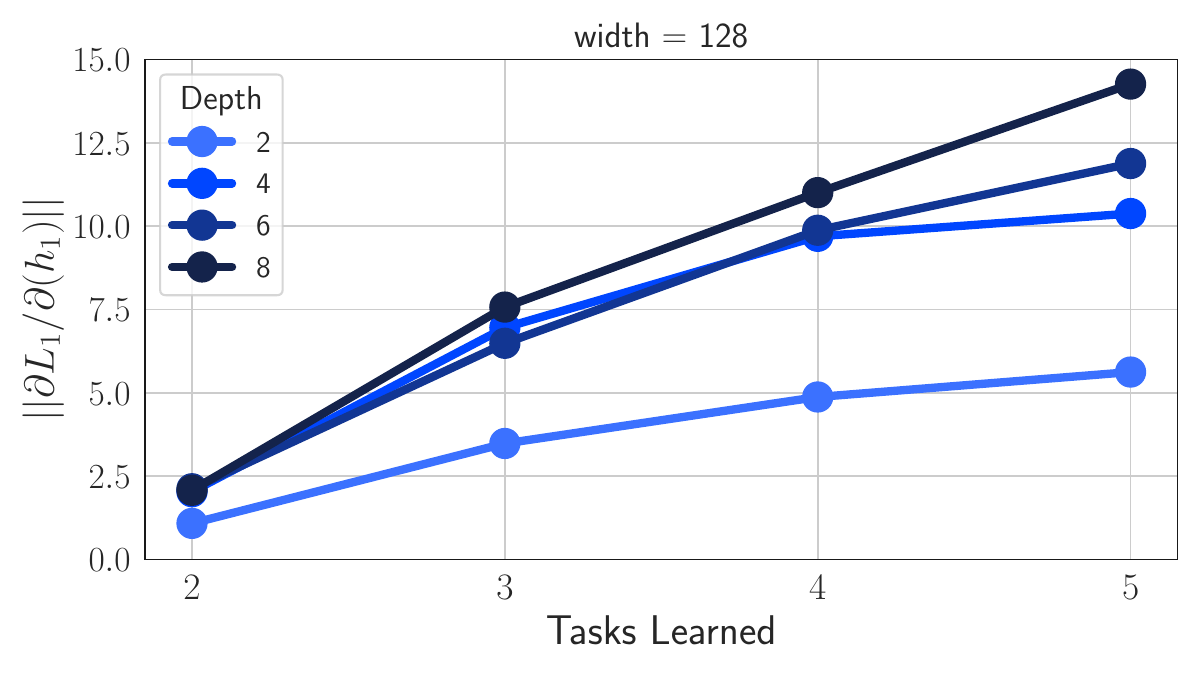}
      \caption{Gradient norm for different depths}
      \label{fig:grad-norm-mnist-depth}
\end{subfigure}\hfill
\begin{subfigure}{.45\linewidth}
      \centering
      \includegraphics[width=.95\linewidth]{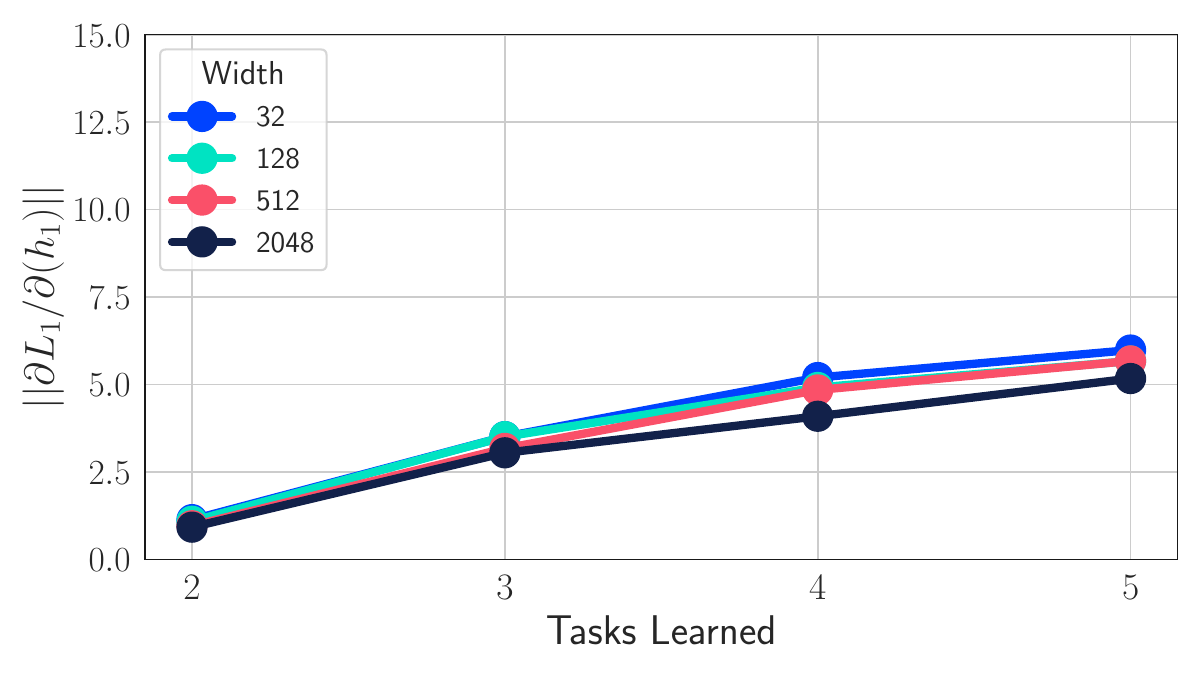}
      \caption{Gradient norm for different widths}
      \label{fig:grad-norm-mnist-width}
\end{subfigure}\hfill
\caption{MLP on Rotated MNIST: The gradient norm in the first layer (i.e.,$\norm{\frac{\partial L_1}{\partial h_{1}}}$)  grows faster in deeper networks (a) while this is not the case when the width increases (b).}
\label{fig:grad-norm}
\end{figure*}

Figure~\ref{fig:grad-norm-mnist-depth} shows the norm of the gradients of layer 1 for different MLP depths. It can be seen from the figure that the gradient norm on the earlier layers increases with the depth. For example, when the network is trained for task 5, the gradient norm on layer 1 for an 8-layer network is almost three times as that of the 2-layer network. In contrast, as depicted in Figure~\ref{fig:grad-norm-mnist-width}, increasing the width has a minimal or even decreasing effect on gradient norm. We note here that the deep learning community has developed numerous strategies to avoid exploding gradients~\citep{goodfellow2016deep}, and that extensively studying those is not the purpose here. We use the exploding gradient analysis to understand the negative effect of depths in our experiments.\vspace{6mm}\newline

In conclusion, we have observed that wider networks have sparser gradients with more orthogonal gradients across tasks. In addition, the training dynamics of wide networks become more similar to the lazy training regime. Finally, the gradient norm in wide and shallow models does not increase as fast as in deeper and thinner models.

\begin{table}[t!]
\centering
\caption{Comparing shallow and wide vs. deep and thin MLP models on Rotated MNIST. Each row in a subdivision of the table has roughly a comparable number of parameters. Shallower and wider networks have higher average accuracy and smaller forgetting than deeper and thinner networks.}
\label{tab:tradeoff-mnist}
\resizebox{\linewidth}{!}{%
\begin{tabular}{@{}lccccl@{}}
\toprule
\textbf{Width} & \textbf{Depth} & \multicolumn{1}{l}{\textbf{Parameters}} & \textbf{\begin{tabular}[c]{@{}c@{}}Average\\ Accuracy\end{tabular}} & \textbf{\begin{tabular}[c]{@{}c@{}}Average\\ Forgetting\end{tabular}} & \multicolumn{1}{c}{\textbf{\begin{tabular}[c]{@{}c@{}}Joint\\ Accuracy\end{tabular}}} \\ \midrule
128 & 8 & 217.35 K & 68.9 $\pm 1.07$ & 35.4 $\pm 1.34$ & 94.1 $\pm 0.73$ \\
256 & 2 & 269.32 K & $\bf{71.1 \pm 0.43}$ & $\bf{31.4 \pm 0.48}$ & 93.9 $\pm 0.65$ \\ \midrule
256 & 8 & 664.08 K & 70.4 $\pm 0.61$ & 32.1 $\pm 0.75$ & 94.78 $\pm 0.67$ \\
512 & 2 & 669.70 K & $\bf{72.6 \pm 0.27}$ & $\bf{29.6 \pm 0.36}$ & 94.08 $\pm 0.77$ \\ \bottomrule
\end{tabular}%
}
\vspace{1mm}
\end{table}

\section{Additional Experiments}\label{sec:additional-results}
This section provides further empirical results on the role of width and depth in continual learning. More specifically, fixing the number of parameters, the wider and shallower models, outperform deeper and thinner ones. Moreover, we show that the benefits of increasing the width can be complementary to the benefits of algorithms.

\vspace{2mm}
\subsection{Width versus Depth}
We have already seen in Figure~\ref{fig:intro-width} and~\ref{fig:intro-depth} that increasing width is beneficial while increasing depth has no such effect. However, it is interesting to answer the following question in a more direct manner: \emph{Given a fixed budget on the total number of parameters, how would a wider and shallower model perform compared to a thinner and deeper one?}\vspace{3mm}

To this end, we compare several architectures on both Rotated MNIST and Split CIFAR-100 benchmarks. Table~\ref{tab:tradeoff-mnist} and~\ref{tab:tradeoff-cifar} show that on both benchmarks, wider and shallower models outperform their thinner and deeper counterparts. Note that to make sure that the number of parameters are roughly the same, we have added some new models (e.g., MLP with a width of 256).

\subsection{Interacting with CL Algorithms}\label{sec:additional-algorithms}

We have seen that increasing width automatically reduces forgetting without the need to employ any continual learning algorithm. One may wonder if the effect of width on performance remains additive even when a specialized CL algorithm is employed. 

\begin{table}[t!]
\centering
\caption{Comparing shallow and wide vs. deep and thin WRN models on Split CIFAR-100. Each row in a subdivision of the table has roughly a comparable number of parameters. Shallower and wider networks have higher average accuracy and smaller forgetting than deeper and thinner networks.}
\label{tab:tradeoff-cifar}
\resizebox{\linewidth}{!}{%
\begin{tabular}{@{}lccccc@{}}
\toprule
\textbf{Depth} & \textbf{Width} & \textbf{Params} & \textbf{\begin{tabular}[c]{@{}c@{}}Average\\ Accuracy\end{tabular}} & \textbf{\begin{tabular}[c]{@{}c@{}}Average\\ Forgetting\end{tabular}} & \textbf{\begin{tabular}[c]{@{}c@{}}Joint\\ Accuracy\end{tabular}} \\ \midrule
10 & 4 & 1.69 M & $\bf{53.8 \pm 2.74}$ & $\bf{33.8 \pm 2.16}$ & 83.4 $\pm 0.64$ \\
28 & 2 & 1.61 M & 46.6 $\pm 2.56$ & 37.1 $\pm 2.47$ & 83.6 $\pm 0.54$ \\ \midrule
10 & 8 & 3.72 M & $\bf{59.7 \pm 2.33}$ & $\bf{29.4 \pm 2.52}$ & 84.8 $\pm 0.49$ \\
16 & 4 & 3.24 M & 50.1 $\pm 2.59$ & 37.0 $\pm 2.77$ & 85.1 $\pm 0.45$ \\
28 & 3 & 3.58 M & 49.4  $\pm 1.82$ & 36.2 $\pm 1.98$ & 84.7 $\pm 0.92$ \\ \bottomrule
\end{tabular}%
}
\vspace{-1mm}
\end{table}

\begin{figure}[t!]
\centering
\begin{subfigure}{.92\linewidth}
      \centering
      \includegraphics[width=.9\linewidth]{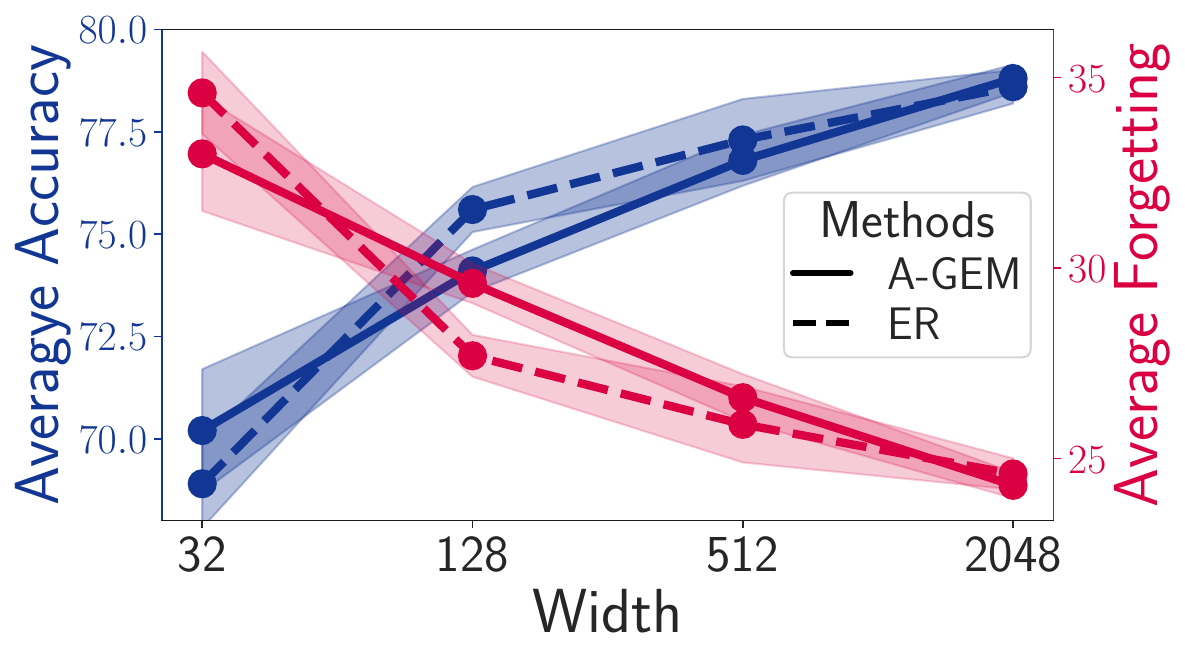}
      \caption{MLP on Rotated MNIST}
      \label{fig:other-methods-mnist}
\end{subfigure}\hfill
\begin{subfigure}{.92\linewidth}
      \centering
      \includegraphics[width=.9\linewidth]{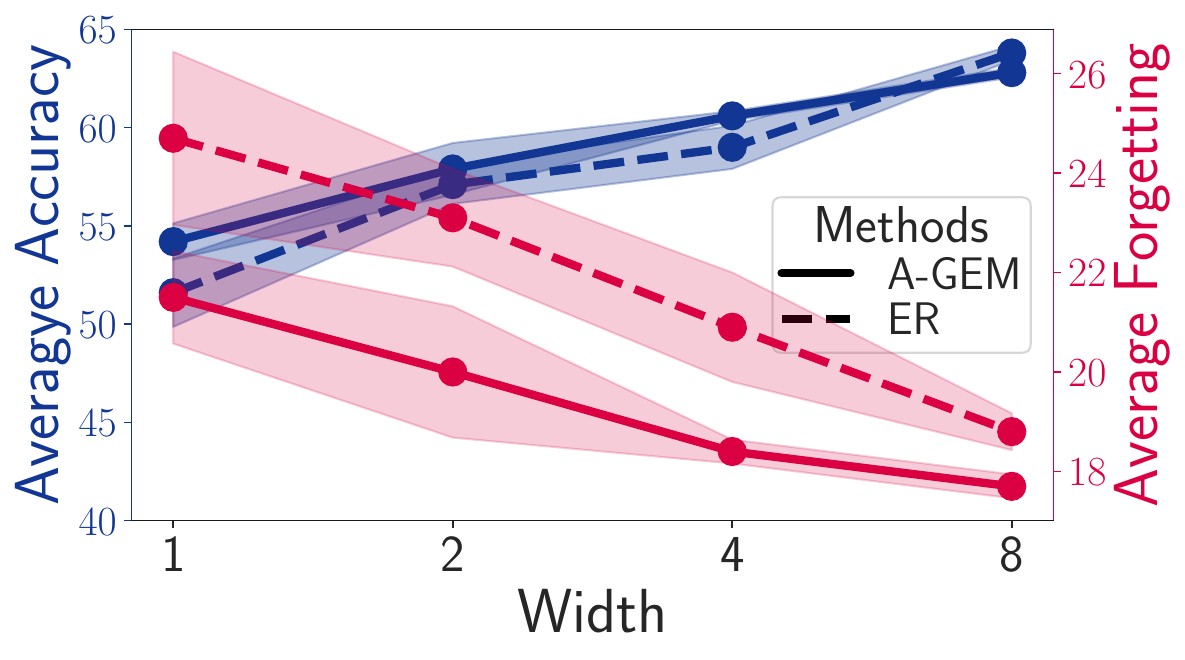}
      \caption{WRN on Split CIFAR-100}
      \label{fig:other-methods-cifar}
\end{subfigure}\hfill
\caption{Average accuracy (blue) and forgetting (red) of CL methods for varying widths. Similar to the naive fine-tuning case, in the presence of other CL algorithms, wider networks perform better.}
\label{fig:other-methods}
\end{figure}

In Figure~\ref{fig:other-methods}, we compare two replay-based methods, ER~\citep{riemer2018learning} and A-GEM~\citep{chaudhry2018efficient}, for various width factors using both MLPs and WideResNets. We use the replay buffer of sizes 125 (MNIST) and 100 (CIFAR-100) for both ER and A-GEM. The figure shows that, similar to vanilla fine-tuning, by increasing width, average accuracy improves, and forgetting gets reduced across the board. Hence, we conclude that the benefit of increasing the width is additive to the benefits brought by continual learning algorithms. 

In Appendix~\ref{apx:additional_algs_mnist}, we provide additional results for more algorithms such as Mode Connectivity SGD~\citep{mirzadeh2021linear}, LwF~\citep{Li2018LearningWF}, and EWC~\citep{EWC} on Rotated MNIST. We refer the reader to ~\citep{Mirzadeh2022ArchitectureMI} for additional results on other benchmarks and architectures such as CNNs, ResNets~\citep{he2016deep}, and Vision Transformers~\citep{dosovitskiy2020image}.

\section{Discussion and Conclusion}\label{sec:discussion}
\looseness=-1 In this paper, we studied a previously unexplored connection between the width of the network and catastrophic forgetting (CF)---an important bottleneck for continual learning (CL) systems. Through different experiments, we hypothesize that high orthogonality between the gradients of different tasks induced by the wider networks, increased gradient sparsity, and a lazy training regime, i.e., the decreased distance of updated parameters from their initialization are the main factors alleviating catastrophic forgetting. We highlighted the contrasts between over-parametrization with increasing width to increasing depth and argued that increasing depth might have negative consequences for forgetting while acknowledging its importance for learning representations. 

The scope of current work is to show a surprising and, at the same time, very intuitive correlation between width and forgetting. One should be cautious of utilizing increasing width to tackle catastrophic forgetting as it is undoubtedly an inefficient decision in terms of computation time, energy and cost. By showing the connection between width and CF, our purpose here is to enhance our basic understanding of how over-parametrization, both in terms of width and depth, interacts with catastrophic forgetting. The extent of over-parametrization should be an important factor when designing continual learning benchmarks.

We hope this work acts as a stepping stone for studying the interaction between the inherent properties of neural networks and catastrophic forgetting. In addition to varying architecture sizes, modern neural networks employ several training tricks that are not well studied with respect to catastrophic forgetting. Understanding these effects, such as activation functions, normalization schemes, skip connections, etc., remains an interesting future work. Very recently,~\citet{Mirzadeh2022ArchitectureMI} have studied the significance of architectures in continual learning. They have extended our main results for other architectures (e.g., Vision Transformers~\citep{dosovitskiy2020image}) on large-scale benchmarks. In addition, they show that the role of architecture in continual learning can be as important as the algorithms, which suggests the intersection of continual learning and neural network architecture is an interesting and under-explored research direction.

\section*{Acknowledgments}
The authors are grateful to Amal Rannen-Triki, Timothy Nguyen, Hooman Shahrokhi, and Anonymous Reviewers for their
valuable comments and feedback on this work.

\bibliography{refs}
\bibliographystyle{apalike}

\clearpage
\newpage

\appendix
\onecolumn
\icmltitle{Wide Neural Networks Forget Less Catastrophically: \\
Supplementary Material}
\setcounter{section}{0}

\section{Experimental Setup Details}\label{apx:experimental-setup}
In this section, we discuss our experimental setup, including the design choices for benchmarks, architectures, and hyper-parameters.

\subsection{Design Choices}
\subsubsection*{Benchmarks} We have chosen the rotated MNIST benchmark since it is a common benchmark in the research literature and, unlike the permuted MNIST, has a meaningful shift across tasks. The number of 5 tasks on MNIST is also typical in the literature~\citep{farajtabar2020orthogonal,mirzadeh2020understanding}. While the rotation of 10 degrees per task is more common in the literature, we have increased the rotation to 22.5 degrees per task make the benchmark more difficult and the performance gap between different models become more visible~\citep{mirzadeh2021linear}. 

While the rotated MNIST benchmark is domain-incremental, we have chosen the split CIFAR-100, which is a common task-incremental benchmark in the literature. Moreover, the split CIFAR-100 has 20 tasks, which helps support our arguments when the number of tasks increases. Our goal was to show our results in various setups such as domain versus task incremental, small versus large number of tasks, and single-head versus multi-head learning.

\subsubsection*{Architectures}
On the MNIST benchmark, MLP is a popular architecture in the literature~\citep{Wortsman2020SupermasksIS,EWC}. However, on the split CIFAR-100, ResNet-18 is a more popular choice~\citep{Chaudhry2019HAL,mirzadeh2021linear}. However, WideResNets (WRN) have two essential properties that are more suitable for this study: first, they share similar architectural properties with ResNets (e.g., skip-connections, batch normalization layers, etc.), and second, they have the nice property of ease of scaling both width and depth.  

We note that for the experiments that involved increasing depth, we have not chosen the thinnest models as a starting point. For instance, in Fig.~\ref{fig:intro-depth}, we have used MLP with a width of 128 and WideResNet with a width of 2. The reason is to make sure that the models are not too small, so by increasing the depth, the number of parameters increases in a meaningful manner.

\subsection{Hyperparameters}
For all experiments, we use a grid to ensure that one specific set of hyper-parameters is not favoring a specific model. Moreover, the training regime plays an important role in continual learning performance~\citep{mirzadeh2020understanding}. 
\begin{enumerate}
    \item learning rate: [0.001, 0.01 (MNIST), 0.05 (CIFAR), 0.1]
    \item batch size: [16, 32 (CIFAR), 64 (MLP)]
    \item SGD momentum: [0.0 (MNIST), 0.8 (CIFAR)]
    \item weight decay: [0.0 (MNIST), 0.0001 (CIFAR)]
\end{enumerate}

For each set of hyper-parameters, we run the model using 5 different seeds for random initialization of the models. Moreover, to ensure that the specific ordering of tasks in the split CIFAR-100 is not playing a role in the results, we also use 3 additional seeds to shuffle the ordering of tasks.

\section{Additional Results}
\label{sec:appendix-additional-results}

\subsection{Evolution of the Average Accuracy Throughout the Learning Experience}

\begin{figure*}[h!]
\centering
\begin{subfigure}{.44\textwidth}
      \centering
      \includegraphics[width=.98\linewidth]{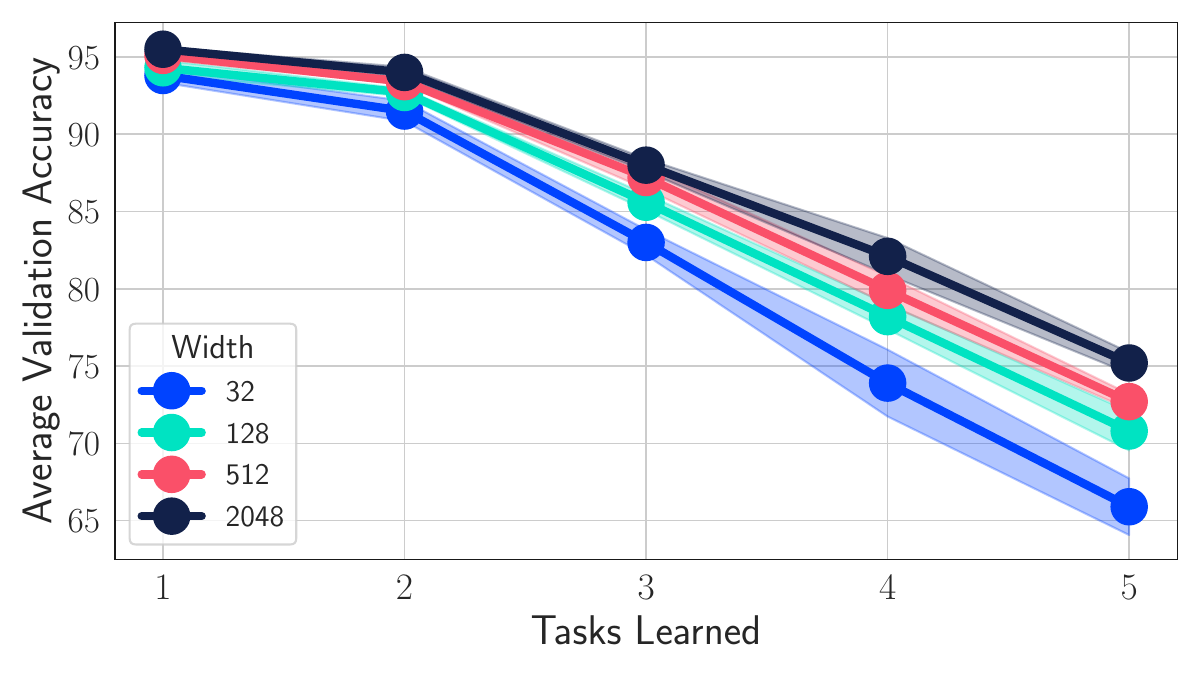}
      \caption{MLP on Rotated MNIST}
      \label{fig:avg-accs-mnist}
\end{subfigure}\hfill
\begin{subfigure}{.44\textwidth}
      \centering
      \includegraphics[width=.99\linewidth]{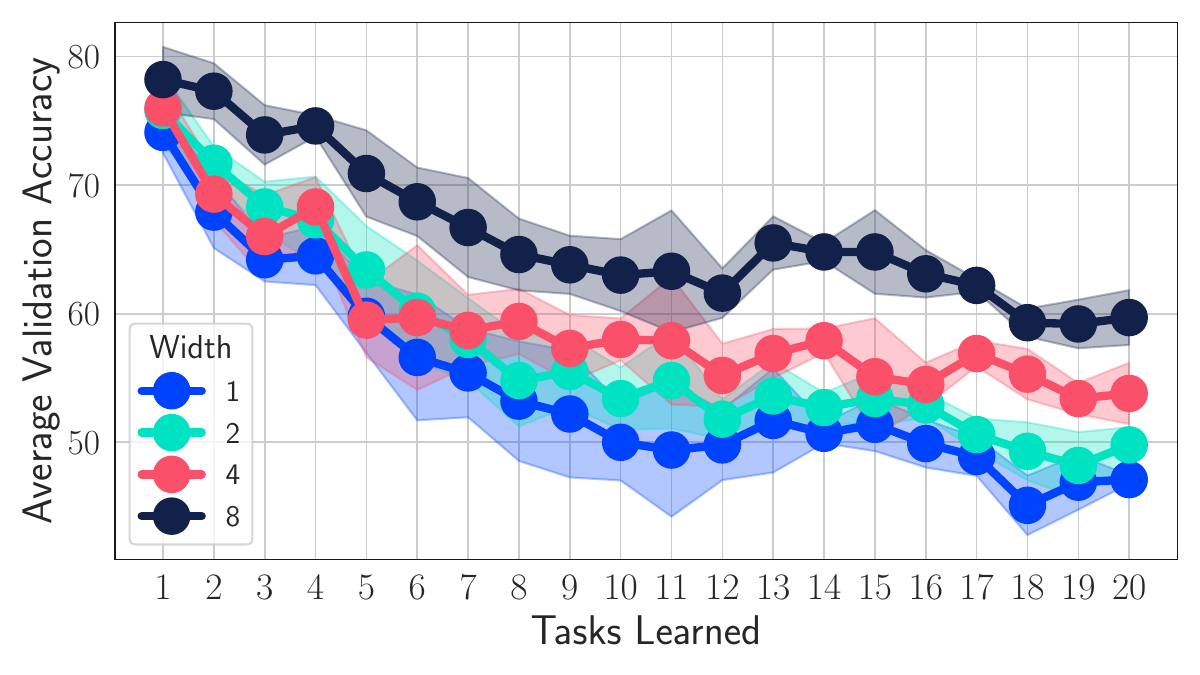}
      \caption{WideResNet on Split CIFAR-100}
      \label{fig:avg-accs-cifar}
\end{subfigure}\hfill
\caption{Evolution of the average accuracy during the continual learning experience: while the gap is not initially large, it grows as the number of tasks increases.}
\label{fig:avg-accs}
\end{figure*}

While tables~\ref{tab:results-mnist} and~\ref{tab:results-cifar} provide the ``final'' average accuracy across different widths, it will be interesting to also compare the ``evolution'' of the average accuracy in throughout of learning experience. Fig.~\ref{fig:avg-accs-mnist} and~\ref{fig:avg-accs-cifar} show this during the learning.

There are a few interesting observations we can make from these results. First, the wider models perform consistently better during the learning experience. Second, the performance gap between wide and thin models is not large initially; however, as learning progresses, the gap increases. Finally, it shows that the decrease in the average accuracy due to forgetting still happens for wide models.  This points out to our discussion in Sec.~\ref{sec:discussion} that even though increasing width has beneficial effects, it is not the ultimate solution to continual learning. However, it will be interesting to understand the reasons behind these beneficial effects and use them in future works to design better continual learning algorithms, and this is the main objective of this study. 

We note that while wider models learn task 1 with higher accuracy, the gap between wide and thin models grows as the number of tasks increases. For instance, in Fig.~\ref{fig:avg-accs-cifar}, the accuracy of WRN-10-8 on task 1 is roughly 4\% higher than WRN-10-1. However, when the learning ends, the accuracy gap between WRN-10-8 and WRN-10-1 grows to more than 12\%. While part of this gap can be explained by the fact that WRN-10-8 is a better learner, we note that the majority of the gap comes from the fact that WRN-10-8 has substantially smaller forgetting, as shown in Table.~\ref{tab:results-cifar}.

\subsection{Varying Width}
In Claim~\ref{claim:fgt} in Sec.~\ref{sec:analysis}, we have theoretically studied the importance of width for a simple case. Here, we provide additional empirical evidence for the Rotated MNIST benchmark where we use a 3-layer MLP model and measure continual learning metrics for different widths (64, 256, 1024) at each layer.

As shown in Table.~\ref{tab:varying width}, we show that the width of the first layer plays a more important role compared to the second and third layers, which is consistent with our analysis. While the models that have a width of 1024 in their first layer have more parameters, we note that we have already discussed in Sec.~\ref{sec:additional-results} that simply increasing the number of parameters is not necessarily helpful for reducing the forgetting.

\begin{table}[h!]
\centering
\caption{Performance metrics for different 3-layer MLP models on Rotated MNIST.}
\label{tab:varying width}
\resizebox{0.6\textwidth}{!}{%
\begin{tabular}{@{}ccccccc@{}}
\toprule
\textbf{L1} & \textbf{L2} & \textbf{L3} & \textbf{Params} & \textbf{\begin{tabular}[c]{@{}c@{}}Average\\ Accuracy\end{tabular}} & \textbf{\begin{tabular}[c]{@{}c@{}}Average\\ Forgetting\end{tabular}} & \textbf{\begin{tabular}[c]{@{}c@{}}Learning\\ Accuracy\end{tabular}} \\ \midrule
64 & 64 & 64 & 59.2 K & 67.9 $\pm 0.28$ & 37.2 $\pm 0.56$ & 97.2 $\pm 0.37$ \\
64 & 64 & 256 & 73.6 K & 68.5 $\pm 0.99$ & 37.1 $\pm 0.18$ & 97.5 $\pm 0.67$ \\
64 & 64 & 1024 & 131.2 K & 68.1 $\pm 0.64$ & 37.2 $\pm 0.82$ & 97.8 $\pm 0.28$ \\
64 & 256 & 64 & 84.0 K & 67.7 $\pm 0.58$ & 37.6 $\pm 0.79$ & 97.7 $\pm 0.29$ \\
64 & 1024 & 64 & 183.1 K & 68.2 $\pm 0.55$ & 36.9 $\pm 0.72$ & 97.8 $\pm 0.31$ \\ \midrule
256 & 256 & 64 & 283.8 K & 71.9 $\pm 0.61$ & 32.8 $\pm 0.84$ & 97.8 $\pm 0.28$ \\
256 & 256 & 256 & 335.1 K & 71.8 $\pm 0.54$ & 32.9 $\pm 0.63$ & 98.1 $\pm 0.33$ \\
256 & 256 & 1024 & 540.2 K & 71.4 $\pm 0.68$ & 33.6 $\pm 0.83$ & 98.0 $\pm 0.37$ \\ \midrule
1024 & 64 & 64 & 874.2 K & 73.1 $\pm 0.94$ & 29.8 $\pm 1.10$ & 97.1 $\pm 0.72$ \\
1024 & 64 & 1024 & 946.2 K & 73.9 $\pm 0.69$ & 29.4 $\pm 0.76$ & 97.4 $\pm 0.52$ \\
1024 & 1024 & 64 & 1.92 M & 74.3 $\pm 0.40$ & 28.5 $\pm 0.51$ & 97.1 $\pm 0.67$ \\
1024 & 1024 & 256 & 2.12 M & 74.4 $\pm 0.35$ & 28.4 $\pm 0.50$ & 97.1 $\pm 0.66$ \\
1024 & 1024 & 1024 & 2.91 M & 75.1 $\pm 0.41$ & 27.8 $\pm 0.52$ & 97.3 $\pm 0.60$ \\ \bottomrule
\end{tabular}%
}
\end{table}

\subsection{Lazy Training Regime in Deeper Models}
\begin{figure*}[h!]
\centering
\begin{subfigure}{.44\textwidth}
      \centering
      \includegraphics[width=.99\linewidth]{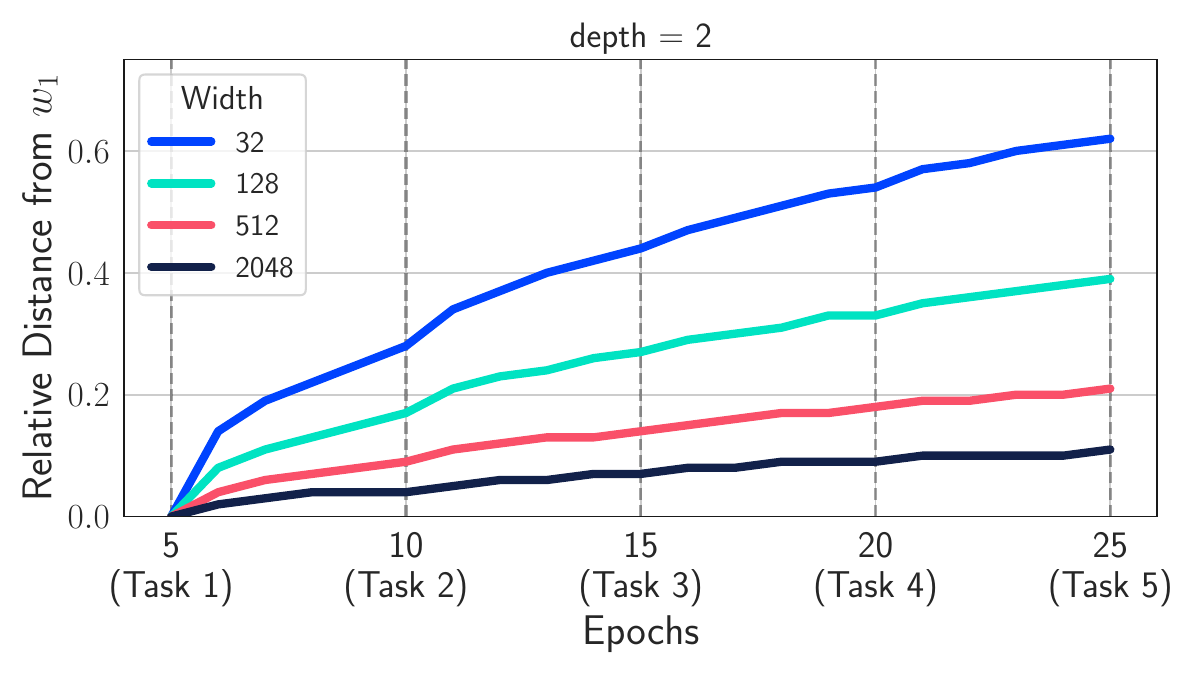}
      \caption{MLP on Rotated MNIST (different widths)}
      \label{fig:lazy-supp-mnist}
\end{subfigure}\hfill
\begin{subfigure}{.44\textwidth}
      \centering
      \includegraphics[width=.99\linewidth]{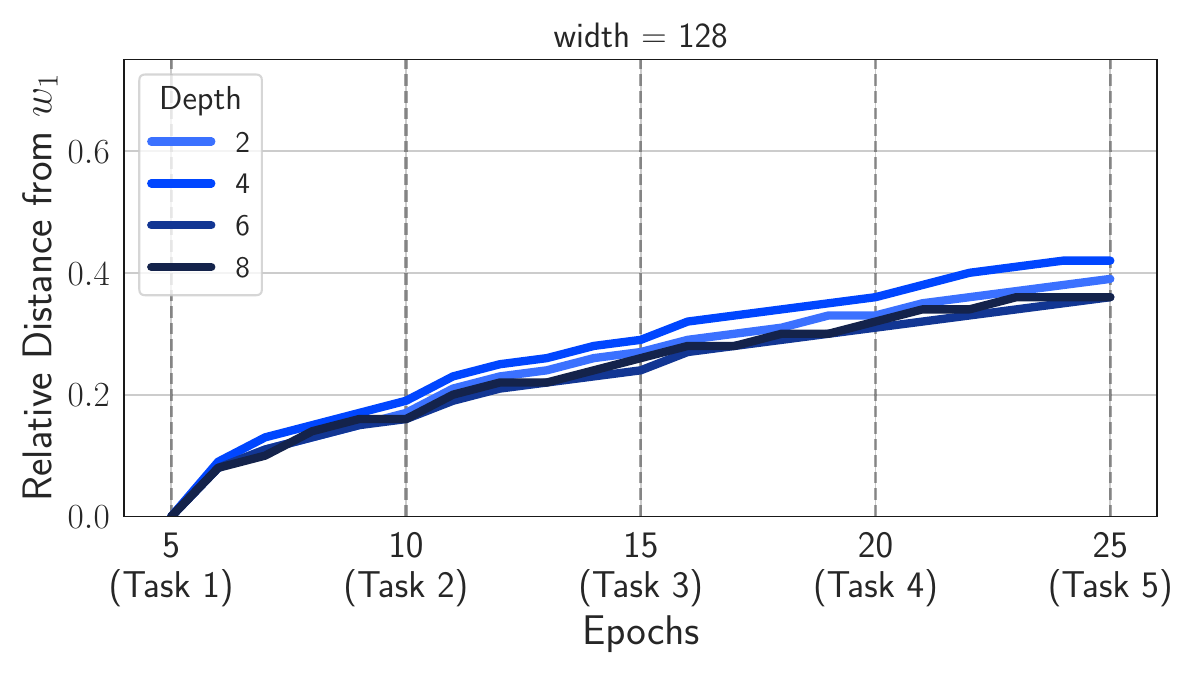}
      \caption{MLP on Rotated MNIST (different depths)}
      \label{fig:lazy-supp-mnist-deep}
\end{subfigure}\hfill
\caption{The distance between $w_1^*$, solution of task 1, and $w_t^*$, solution of the $t$-th task, for $t > 1$ over the course of training. Wider networks remain closer to the solution of the first task, but deeper models do not show the same property.}
\label{fig:lazy-supp}
\end{figure*}

In Sec.~\ref{sec:lazy-training-regime}, we have discussed the role of parameter changes in the network. In Fig.~\ref{fig:lazy-mnist}, we have provided empirical evidence each parameter of wide models changes less during the learning. However, one important argument could be that the norm of parameters increases as the number of parameters increases, and this might be the reason behind the result that the relative distance of wider models is smaller. However, we show in Fig.~\ref{fig:lazy-supp-mnist-deep} that this is not the case, and the relative distance for deeper models does change as depth increases. Fig.~\ref{fig:lazy-supp-mnist} is similar to Fig.~\ref{fig:lazy-mnist} and was put here for the ease comparison. 

\subsection{Detailed Results for Other Methods}
In Fig.~\ref{fig:other-methods} of Sec.~\ref{sec:additional-results}, we have shown the average forgetting and accuracy of the A-GEM and Experience Replay(ER) algorithms to illustrate that the gain due to increasing width also happens in the presence of other popular continual learning algorithms. Tables~\ref{tab:others-mnist} and~\ref{tab:others-cifar} show the exact numbers for the mentioned figures, and we can see that by increasing width, the average accuracy increases and average forgetting decreases consistently.

Moreover, we have included the data from tables~\ref{tab:results-mnist} and~\ref{tab:results-cifar} for ease of comparison. An interesting observation from these results could be that adding replay buffer or gradient manipulation by the A-GEM algorithm has an additive effect to the gain due to width.

\begin{table*}[h!]
\parbox{.48\textwidth}{
\caption{Rotated MNIST: the gain due to width for different CL algorithms.}
\resizebox{0.99\linewidth}{!}{%
\begin{tabular}{@{}cccc@{}}
\toprule
\textbf{Width} & \textbf{Method} & \textbf{Average Accuracy} & \textbf{Average Forgetting} \\ \midrule
32 & Naive & 65.9 $\pm 1.00 $ & 36.9 $\pm 1.27$ \\
128 & Naive & 70.8 $\pm 0.68$ & 31.5 $\pm 0.92$\\
512 & Naive & 72.6 $\pm 0.27$ & 29.6 $\pm 0.36$ \\
2048 & Naive & 75.2 $\pm 0.34$ & 26.7 $\pm 0.50$ \\ \midrule

32 & ER & 68.9 $\pm 1.08$ & 34.6 $\pm 1.29$ \\
128 & ER & 75.6 $\pm 0.55$ & 27.7 $\pm 0.72$ \\
512 & ER & 77.3 $\pm 1.00$ & 25.9 $\pm 1.28$ \\
2048 & ER & 78.6 $\pm 0.41$ & 24.6 $\pm 0.45$ \\ \midrule

32 & A-GEM & 70.2 $\pm 1.50$ & 33.0 $\pm 1.91$ \\
128 & A-GEM & 74.1 $\pm 0.52$ & 29.6 $\pm 0.73$ \\
512 & A-GEM & 76.8 $\pm 0.62$ & 26.6 $\pm 0.74$ \\
2048 & A-GEM & 78.8 $\pm 0.34$ & 24.3 $\pm 0.50$ \\ \bottomrule

\end{tabular}%
}
\label{tab:others-mnist}
}
\hfill
\parbox{.48\textwidth}{
\caption{Split CIFAR-100: the gain due to width for different CL Algorithms.}
\resizebox{0.99\linewidth}{!}{%
\begin{tabular}{@{}cccc@{}}
\toprule
\textbf{Width} & \textbf{Method} & \textbf{Average Accuracy} & \textbf{Average Forgetting} \\ \midrule

1 & Naive & 47.1 $\pm 2.60$ & 37.3 $\pm 2.62$\\
2 & Naive & 49.7 $\pm 1.51$ & 34.9 $\pm 1.72$\\
4 & Naive & 53.8 $\pm 2.74$ & 33.8  $\pm 2.16$ \\
8 & Naive & 59.7 $\pm 2.33$ & 29.4 $\pm 2.52$ \\
\midrule
1 & ER & 51.6 $\pm 1.74$ & 24.7 $\pm 0.86$ \\
2 & ER & 57.1 $\pm 0.98$ & 23.1 $\pm 1.74$ \\
4 & ER & 59.0 $\pm 1.10$ & 20.9 $\pm 2.45$ \\
8 & ER & 63.8 $\pm 0.37$ & 18.8 $\pm 0.49$ \\
\midrule
1 & A-GEM & 54.2 $\pm 0.93$ & 21.5 $\pm 0.03$ \\
2 & A-GEM & 57.9 $\pm 1.32$ & 20.0 $\pm 1.11$ \\
4 & A-GEM & 60.6 $\pm 0.24$ & 18.4 $\pm 0.79$ \\
8 & A-GEM & 62.8 $\pm 0.24$ & 17.7 $\pm 0.06$ \\ \bottomrule
\end{tabular}%
}
\label{tab:others-cifar}
}
\end{table*}

\subsection{Orthogonality Between Task 2 and Subsequent Tasks}
\begin{figure*}[ht]
\centering
\begin{subfigure}{.48\textwidth}
      \centering
      \includegraphics[width=.98\linewidth]{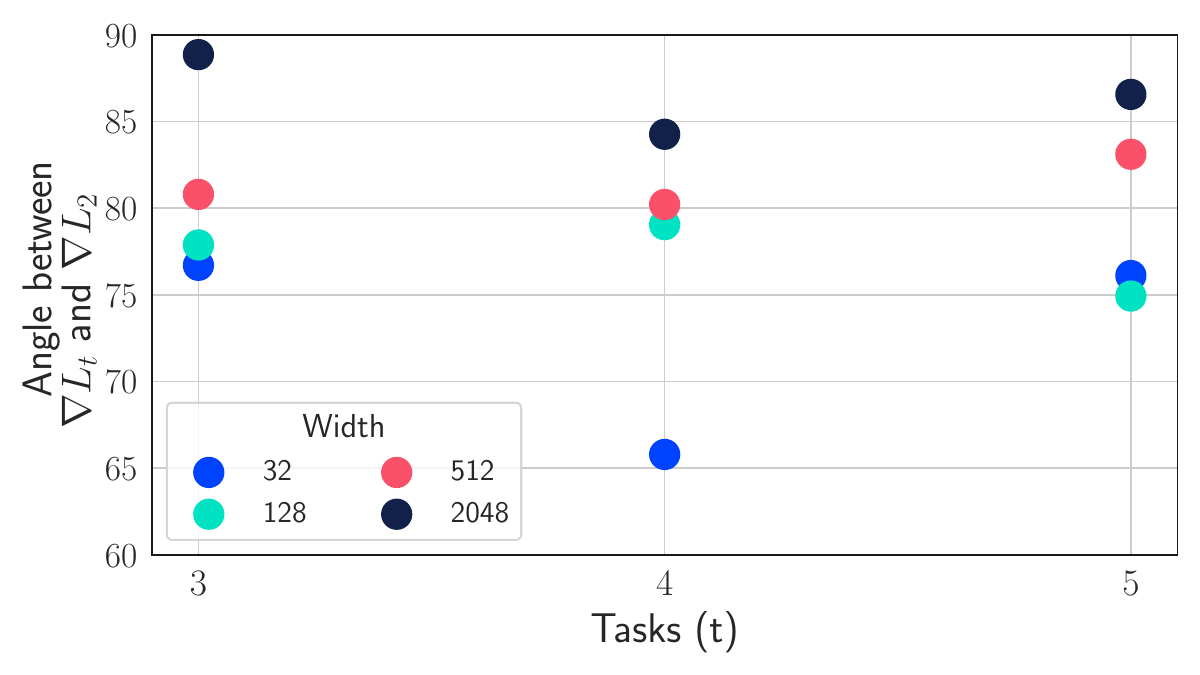}
      \caption{MLP on Rotated MNIST}
      \label{fig:ortho-mnist-t2}
\end{subfigure}\hfill
\begin{subfigure}{.48\textwidth}
      \centering
      \includegraphics[width=.98\linewidth]{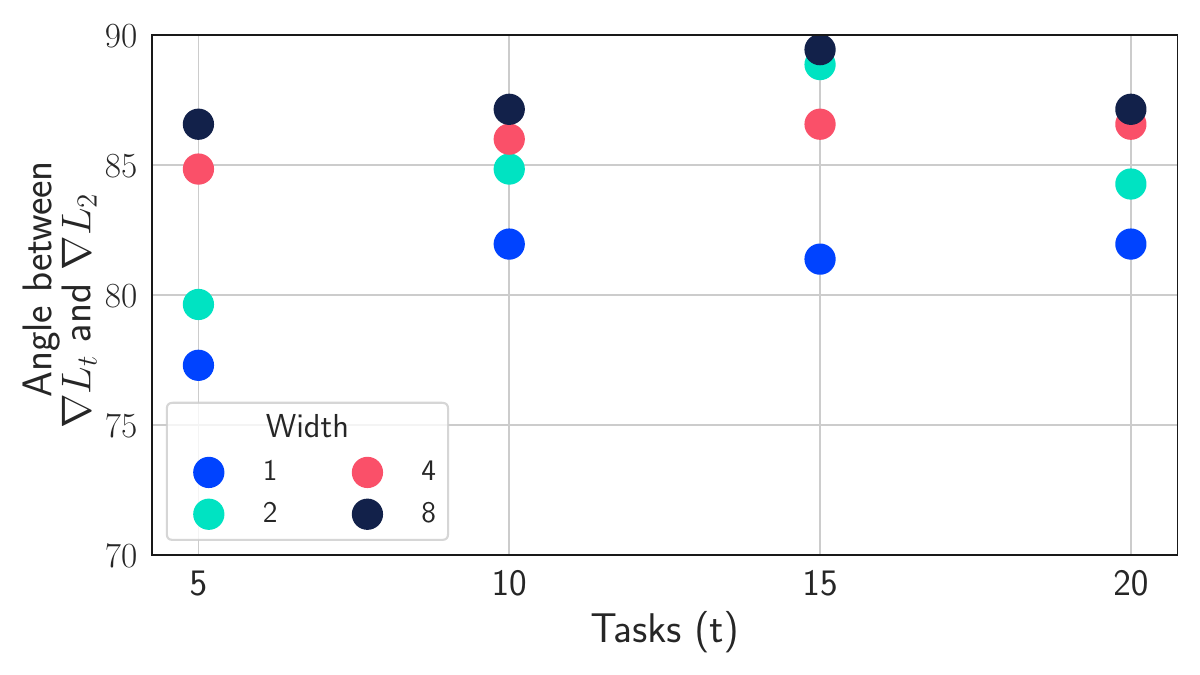}
      \caption{WideResNet on Split CIFAR-100}
      \label{fig:ortho-cifar-t2}
\end{subfigure}\hfill
\caption{The degree between the gradients of task 2 and the gradients of subsequent tasks}
\label{fig:ortho-t2}
\end{figure*}

To show that our conclusion in Section~\ref{sec:analysis} about orthogonality of gradients holds for the next tasks, in this section, we provide a similar figure to Fig.~\ref{fig:ortho} but for task 2. In Fig.~\ref{fig:ortho-t2}, we plot the angle between the gradient vector at the minimum of the second task and that of the subsequent tasks for various width factors. We can see that for wider models, the trend of gradients being orthogonal continues after learning the second task as well.

\subsection{Singular Values of Deeper Models}\label{apx:svs-deeper}
\begin{figure*}[ht]
\centering
\begin{subfigure}{.48\textwidth}
      \centering
      \includegraphics[width=.98\linewidth]{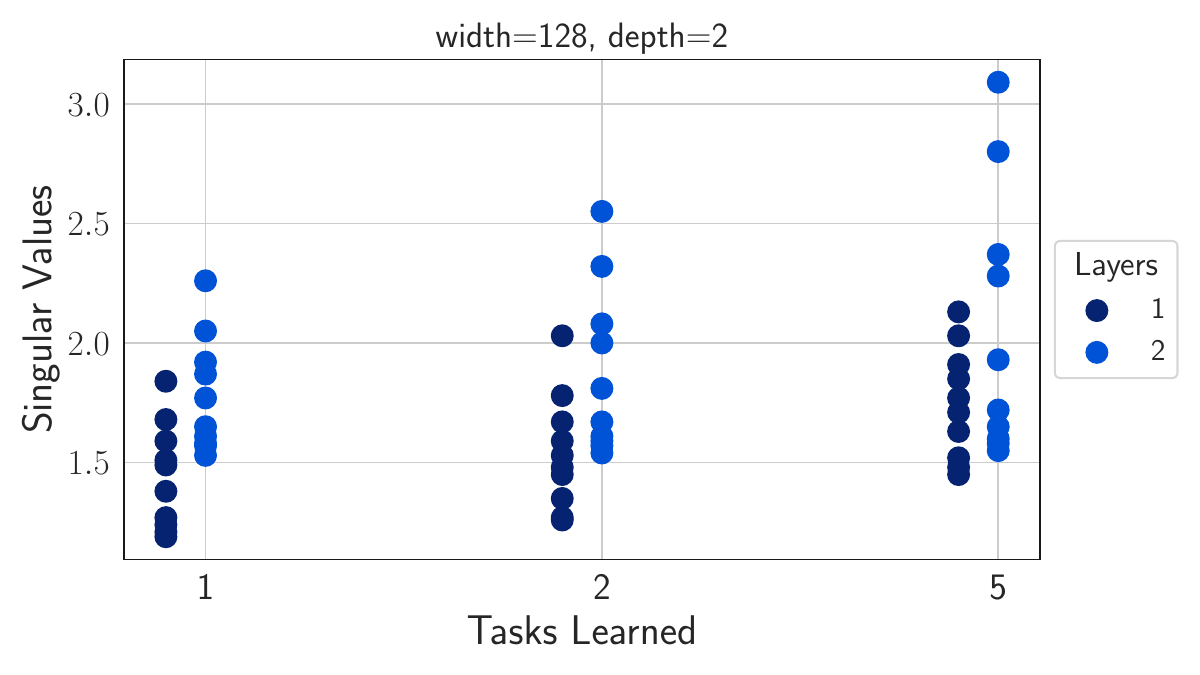}
      \caption{Rotated MNIST: depth = 2}
      \label{fig:sv-d2}
\end{subfigure}\hfill
\begin{subfigure}{.48\textwidth}
      \centering
      \includegraphics[width=.98\linewidth]{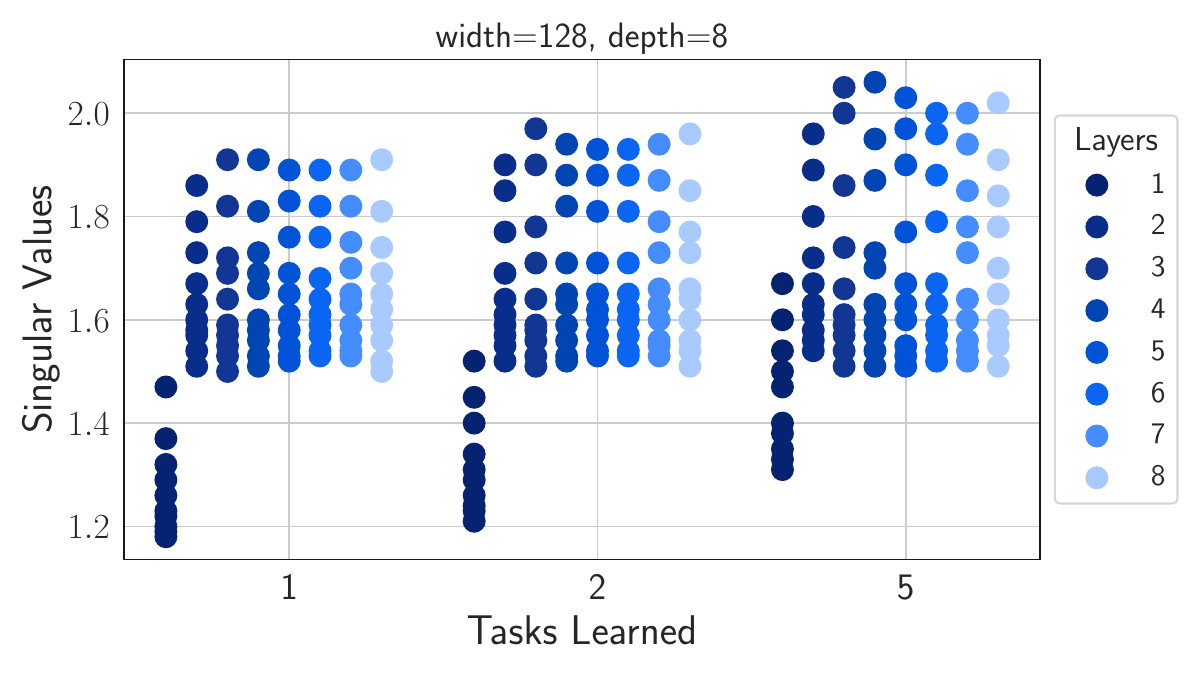}
      \caption{Rotated MNIST: depth = 8}
      \label{fig:sv-d8}
\end{subfigure}\hfill
\caption{Top-10 singular values of each layer during the learning: top singular values are larger than 1.}
\label{fig:sv}
\end{figure*}

In Eq.~\eqref{eq:sv-bound} of Sec.~\ref{sec:depth_cf}, we have discussed that one condition that impacts our analysis, is the singular values of all intermediate layers. Fig.~\ref{fig:sv-d2} shows the singular values for our shallowest model with depth of 2, and Fig.~\ref{fig:sv-d8} shows the values for the deepest model. We can see that the top singular values are all greater than 1.

We note that to make the figures easier to read, we show the top-10 singular values at the end of tasks 1, 2, and 5.

Finally, we would like to provide some clarifications on our analysis in Sec.~\ref{sec:depth_cf}. We note that the bound provided for the gradient norm is unusable for $\Lambda_i > 1$, as it is an upper bound which grows exponentially -- becomes more and more loose -- with $K$ (the number of layers). The purpose of providing the analysis is to potentially point to a scenario when the gradients on the earlier layers could increase with the depth. This \emph{can} happen when the singular values of all the intermediate weight matrices are greater than 1, \emph{i.e.}, $\Lambda_i > 1, \forall i \in \{l, \cdots, K-1\}$. In Fig.~\ref{fig:sv}, we empirically see that this indeed is the case for Rotated MNIST tested with varying MLP depths.

\subsection{Interacting With Additional Algorithms}\label{apx:additional_algs_mnist}
In Sec.~\ref{sec:additional-algorithms}, we have shown that similar to the naive finetuning case, when using A-GEM~\citep{chaudhry2018efficient}, and Experience Replay, wide networks perform better for both Rotated MNIST and Split CIFAR-100 benchmarks.

Here, we extend the result for other algorithms on Rotated MNIST (with 5 tasks). We use Mode Connectivity SGD~(MC-SGD)~\citep{mirzadeh2021linear}, which is the state-of-the-art algorithm in continual learning, in addition to Learning without Forgetting~(LwF)~\citep{Li2018LearningWF} and Elastic Weight Consolidation~(EWC)~\cite{EWC} methods.

\begin{figure*}[t!]
\centering
\begin{subfigure}{.48\textwidth}
      \centering
      \includegraphics[width=.8\linewidth]{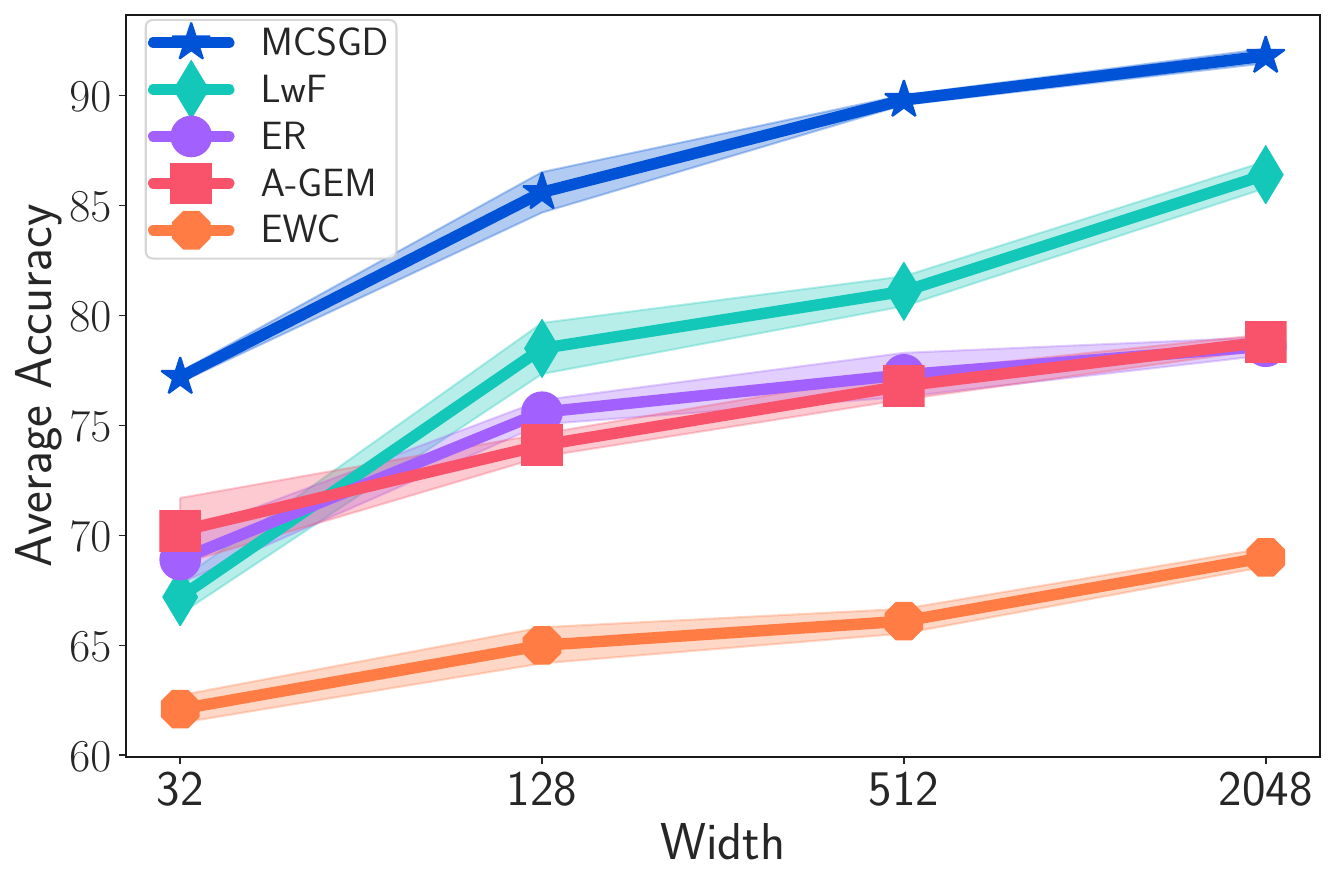}
      \caption{Rotated MNIST: Average Accuracy}
\end{subfigure}\hfill
\begin{subfigure}{.48\textwidth}
      \centering
      \includegraphics[width=.8\linewidth]{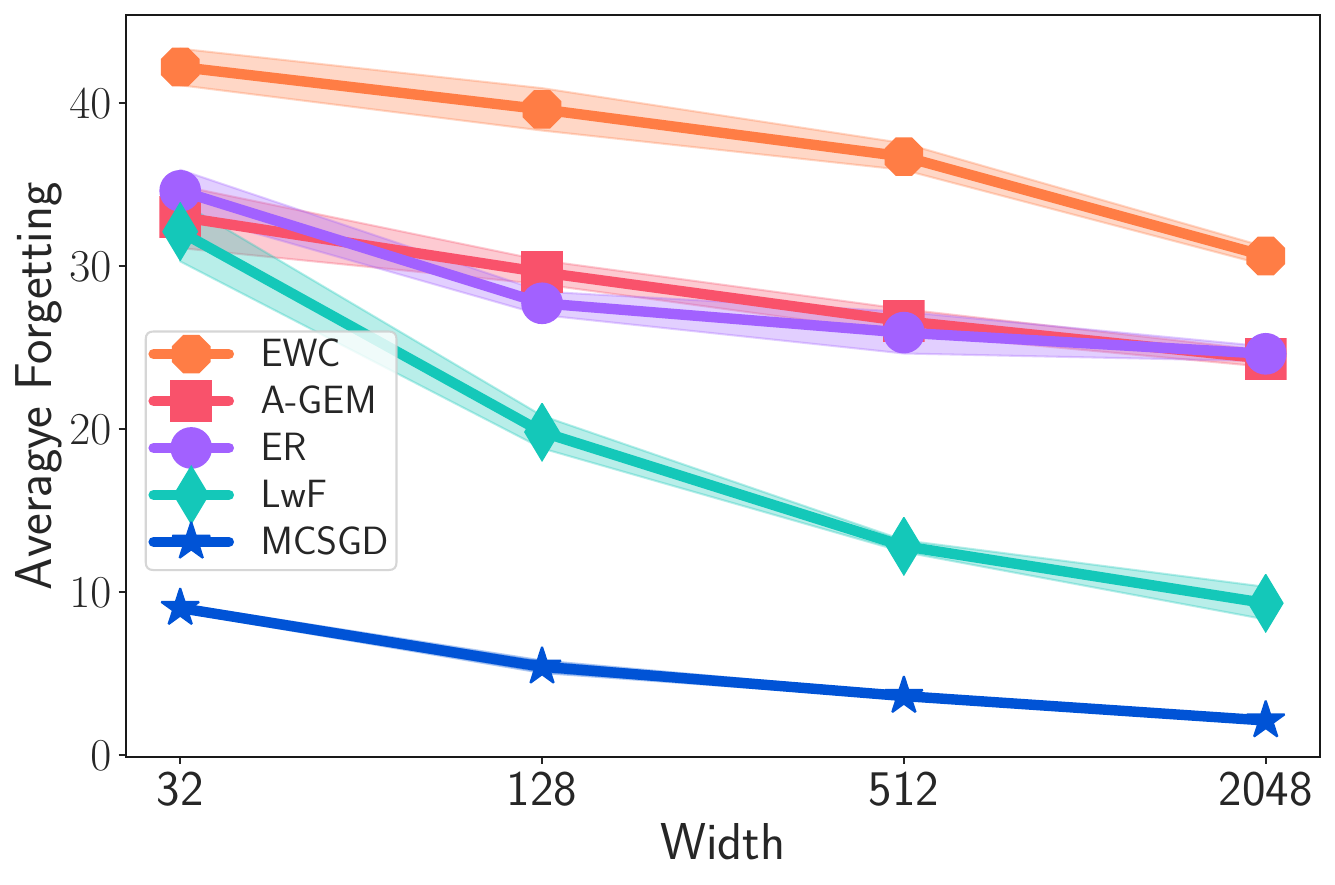}
      \caption{Rotated MNIST: Average Forgetting}
\end{subfigure}\hfill
\caption{Rotated MNIST: Similar to the naive fine-tuning case, in the presence of other CL algorithms, wider networks perform better.}
\label{fig:additional-algs-mnist}
\end{figure*}

Fig.~\ref{fig:additional-algs-mnist} shows that for all algorithms, by increasing the width, the average accuracy increases and average forgetting decreases. However, interestingly, the performance gain varies across algorithms, and methods such as MC-SGD and LwF benefit more from the increased width compared to other algorithms. An interesting future direction is to study this phenomenon in more detail.

\section{Task Construction in Claim~\ref{claim:fgt}}\label{apx:task_construction}

Consider a learning problem where the input space is $\RR^d$ and the output space is $\RR$, and two model classes, $\cF_1,\cF_2\subseteq\{\RR^d\mapsto\RR\}$, where $\cF_1$ is parameterized by $w\in\RR^d$:
\[
\cF_1 := \{f_w: f_w(x) = \langle x, w \rangle, w\in\RR^d\},
\]
and $\cF_2$ is parameterized by $U\in\RR^{h\times d}$ and $v\in\RR^h$:
\[
\cF_2 :=\{f: f_{U,v}(x) = v^\top Ux, U\in\RR^{h\times d}, v\in\RR^h\}.
\]
This means that $\cF_1$ is the set of linear models, and $\cF_2$ is the set of two-layer linear networks. Clearly, we have $\cF_1=\cF_2$, i.e., they have the same capacity, since there is no non-linearity used in $\cF_2$. For any model $f\in\cF_1$ or $\cF_2$, and data point $(x, y)\in \RR^d\times \RR$, we use the quadratic loss function $\ell(f; x, y) = \frac{1}{2}(y-f(x))^2$.

Now consider two tasks with training data points $\{(x_{1,i}, y_{1,i})\}_{i=1}^n$ and $\{(x_{2,i}, y_{2,i})\}_{i=1}^n$. We denote by $X_t = [x_{t, i} \cdots x_{t, n}]^\top\in\RR^{n\times d}$ and $Y_t=[y_{t, i} \cdots y_{t,n}]^\top\in\RR^n$ the data matrix and label vector for task $t$ ($t=1, 2$), respectively.
Define the training loss function:
\[
L_t(f)=\frac{1}{n}\sum_{i=1}^n\ell(f;x_{t,i}, y_{t,i}), \quad t=1,2.
\]

For either model class, the learner first trains the model with data for task 1, and obtains model $f_1$ and then after training ends, the data for task 1 are all removed. The learner then trains the model with data for task 2.
Let $f_2$ be the model that the learner obtains after training on task 2. The \emph{forgetting} metric is defined as
\[
\mathsf{fgt}(f_2, f_1) := L_1(f_2) - L_1(f_1).
\]

We will consider the over-parmeterized regime where $d \gg n$. We know that as long as $X_1X_1^\top$ is invertable, there exist models with zero loss on task 1. In this case, let us assume that the model is sufficiently trained on task 1, i.e., at the end of task 1, we have $L_1(f_1)=0$ for both model classes. Then we train the model on task 2 using gradient descent for $T$ steps with learning rate $\eta$. Let $w^0$ and $w^T$ be the initial and end parameters for model class $\cF_1$ in task 2 training, respectively, and let $(U^0, v^0)$, $(U^T, v^T)$ be the initial and end parameters for model class $\cF_2$ in task 2 training, respectively. Then we have the following claim.

\begin{claim}[Claim~\ref{claim:fgt} restated]
There exist two tasks such that when we train task $2$ using $T$ steps of gradient descent, if we use model class $\cF_1$, the amount of forgetting is strictly zero, i.e.,
\[
    \mathsf{fgt}(f_{w^T}, f_{w^0})=0;
\]
whereas if we use model class $\cF_2$, the amount of forgetting can be positive, i.e., 
\[
\mathsf{fgt}(f_{U^T; v^T}, f_{U^0, v^0}) \ge 0.
\]
In particular, if $h\le n$, and the matrix $X_1(U^0)^\top\in\RR^{n\times h}$ has rank $h$, and $v^T\neq v^0$, then the forgetting for model class $\cF_2$ is positive, i.e., $\mathsf{fgt}(f_{U^T; v^T}, f_{U^0, v^0}) > 0$.
\end{claim}

\begin{proof}
We construct the two tasks in the following way.
We assume that the input features of the two tasks satisfy 
\[
\langle x_{1,i}, x_{2,j} \rangle = 0 \quad \text{for all  } 1\le i, j \le n,
\]
i.e., $X_1X_2^\top = 0$.
This means that the data points in the two tasks are in two orthogonal subspaces.

When training on task 2, the learner use gradient descent algorithm with respect to the parameters. Here, we rewrite the loss function such that it becomes a function of the parameters. For model class $\cF_1$, we let
\[
L_2^1(w):=\frac{1}{2n} \|Y_2 - X_2 w\|_2^2,
\]
and for function class $\cF_2$, we let
\[
L_2^2(U, v):=\frac{1}{2n} \|Y_2 - X_2 U^\top v\|_2^2.
\]

Consider model class $\cF_1$. With learning rate $\eta$, the parameter update rule for model class $\cF_1$ is
\[
w^{t+1} = w^t - \frac{\eta}{n}(X_2^\top X_2 w^t - X_2^\top Y_2).
\]
Since $X_1X_2^\top = 0$, we know $X_1(w^{t+1}- w^t) = 0$. Let $w^0$ be the model parameter that we have at the beginning of the training of task 2. Then after $T$ steps training on task 2, we have $X_1w^T = X_1w^0$. Thus the forgetting for function class $\cF_1$ is zero, i.e.,
\begin{align}
    \mathsf{fgt}(f_{w^T}, f_{w^0})=0.
\end{align}

Now consider model class $\cF_2$. The parameter update rule is
\begin{align*}
    U^{t+1} &= U^t - \frac{\eta}{n}(v^t(v^t)^\top U^t X_2^\top X_2 - v^t Y_2^\top X_2) \\
    v^{t+1} &= v^t - \frac{\eta}{n}(U^t X_2^\top X_2 (U^t)^\top v^t - U^t X_2^\top Y_2).
\end{align*}
Suppose that the second task starts with $f_1$ with parameters $U^0$ and $v^0$. We assume that task 1 is well trained, so $L_1(U^0, v^0)=0$. Since $X_1^\top X_2=0$, we know that $X_1((U^{t+1})^\top - (U^t)^\top)=0$, thus $X_1(U^t)^\top=X_1(U^0)^\top$ for any $t$. Therefore
\begin{align*}
\mathsf{fgt}(f_{U^T, v^T}, f_{U^0, v^0})&=\frac{1}{2n}\|X_1(U^T)^\top v^T-Y_1\|_2^2\\ &=\frac{1}{2n}\|X_1(U^0)^\top v^T-Y_1\|_2^2 \\
&=\frac{1}{2n}\| X_1(U^0)^\top(v^T-v^0) \|_2^2\ge 0.
\end{align*}
One can also easily observe that if $h\le n$, and the matrix $X_1(U^0)^\top\in\RR^{n\times h}$ has rank $h$, and $v^T\neq v^0$, then $\mathsf{fgt}(f_{U^T, v^T}, f_{U^0, v^0}) > 0$.
\end{proof}

\end{document}